\documentclass[12pt, draftclsnofoot, onecolumn]{IEEEtran}

\usepackage[encapsulated]{CJK}
\usepackage{ucs}
\usepackage[utf8x]{inputenc}
\usepackage[cmex10]{amsmath}
\usepackage{amsmath,amssymb,amscd,bbm,amsthm,mathrsfs,dsfont}
\usepackage{algorithmic,algorithm}
\usepackage{mdwmath}
\usepackage{mdwtab}
\usepackage{bm,upgreek}
\usepackage{cite}
\usepackage{rotating,graphics,psfrag,epsfig}
\usepackage{array}
\usepackage{booktabs}
\usepackage{indentfirst}
\usepackage{subfigure}
\usepackage{lipsum,fancyhdr,lastpage,refcount}
\usepackage{url}
\usepackage{blindtext}
\usepackage[T1]{fontenc}
\usepackage{color}
\usepackage{stfloats}
\usepackage{longtable}

\IEEEoverridecommandlockouts

\let\oldnl\nl
\newcommand{\nonl}{\renewcommand{\nl}{\let\nl\oldnl}}

\begin{document}

\title{\huge Age of Information-Aware Radio Resource Management in Vehicular Networks: A Proactive Deep Reinforcement Learning Perspective}

\author{\IEEEauthorblockN{Xianfu Chen, Celimuge Wu, Tao Chen, Honggang Zhang, Zhi Liu, Yan Zhang, and Mehdi Bennis}

\thanks{X. Chen and T. Chen are with the VTT Technical Research Centre of Finland Ltd, Finland (email: \{xianfu.chen, tao.chen\}@vtt.fi).}

\thanks{C. Wu is with the Graduate School of Informatics and Engineering, University of Electro-Communications, Japan (email: clmg@is.uec.ac.jp).}

\thanks{H. Zhang is with the College of Information Science and Electronic Engineering, Zhejiang University, China (e-mail: honggangzhang@zju.edu.cn).}

\thanks{Z. Liu is with the Department of Mathematical and Systems Engineering, Shizuoka University, Japan (email: liu@ieee.org).}

\thanks{Y. Zhang is with the Department of Informatics, University of Oslo, Norway (e-mail: yanzhang@ieee.org).}

\thanks{M. Bennis is with the Centre for Wireless Communications, University of Oulu, Finland (email: mehdi.bennis@oulu.fi).}

\thanks{This work has been submitted to the IEEE for possible publication. Copyright may be transferred without notice, after which this version may no longer be accessible.}
}

\maketitle

\begin{abstract}

In this paper, we investigate the problem of age of information (AoI)-aware radio resource management for expected long-term performance optimization in a Manhattan grid vehicle-to-vehicle network.
With the observation of global network state at each scheduling slot, the roadside unit (RSU) allocates the frequency bands and schedules packet transmissions for all vehicle user equipment-pairs (VUE-pairs).
We model the stochastic decision-making procedure as a discrete-time single-agent Markov decision process (MDP).
The technical challenges in solving the optimal control policy originate from high spatial mobility and temporally varying traffic information arrivals of the VUE-pairs.
To make the problem solving tractable, we first decompose the original MDP into a series of per-VUE-pair MDPs.
Then we propose a proactive algorithm based on long short-term memory and deep reinforcement learning techniques to address the partial observability and the curse of high dimensionality in local network state space faced by each VUE-pair.
With the proposed algorithm, the RSU makes the optimal frequency band allocation and packet scheduling decision at each scheduling slot in a decentralized way in accordance with the partial observations of the global network state at the VUE-pairs.
Numerical experiments validate the theoretical analysis and demonstrate the significant performance improvements from the proposed algorithm.

\end{abstract}


\section{Introduction}
\label{intr}

Vehicle-to-vehicle (V2V) communication is envisioned as one of the key enablers for spearheading the next generation intelligent transportation systems \cite{Wu18, Camp17, Kuut18, Ahma19, Li17, Li19}.
Yet this ad hoc type of vehicular communication requires intense coordinations among the vehicles in close proximity \cite{Wu18, Amad16}, and the high vehicle mobility makes designing efficient radio resource management (RRM) schemes an extremely challenging problem \cite{Zhen15, Qian17, Xiao19}.
There is a huge body of literature on RRM in V2V communications.
In \cite{Sun16}, Sun et al. proposed a separate resource block and power allocation algorithm for RRM in device-to-device based V2V communications.
In \cite{Yao17}, Yao et al. derived a loss differentiation rate adaptation scheme to meet the stringent delay and reliability requirements for V2V safety communications.
In \cite{Egea16}, Egea-Lopez et al. designed a fair adaptive beaconing rate algorithm for beaconing rate control in inter-vehicular communications.
In \cite{Guo19}, Guo et al. developed a joint spectrum and power allocation algorithm based on the effective capacity theory with the aim of maximizing the sum ergodic capacity of vehicle-to-infrastructure links while guaranteeing the latency violation probability for V2V links.
However, while interesting, most efforts have focused on the instantaneous performance optimization ignoring the network dynamics, such as the temporal and spatial variations in communication quality and traffic information.

A Markov decision process (MDP) framework has been widely applied to model the problem of long-term RRM in time-varying vehicular networks \cite{Ye18}.
In \cite{Liu18}, Liu et al. formulated a latency and reliability \cite{Meh18} constrained transmit power minimization problem, in which the Lyapunov stochastic optimization was leveraged to handle the network dynamics.
The Lyapunov optimization can only construct an approximately optimal solution.
In \cite{Chen18S}, we studied the non-cooperative RRM in vehicular communications from an oblivious game-theoretic perspective and put forward an online reinforcement learning algorithm (RL) to approach the optimal control policy.
In \cite{Zhen16}, Zheng et al. proposed a stochastic learning scheme for delay-optimal virtualized radio resource scheduling in a software-defined vehicular network.
For a more practical vehicular network, the explosion in the state space makes the techniques in \cite{Chen18S, Zhen16} infeasible.
In recent years, deep neural networks have been revisited to deal with the curse of state space explosion \cite{Chen19J, Chen19M, Mnih15}.
For example, in \cite{Ye19}, Ye et al. devised a decentralized RRM mechanism based on deep RL (DRL) for V2V communication systems.
In \cite{Lian19}, Liang et al. proposed a fingerprint-based deep Q-network (DQN) method to solve the RRM problem in vehicular networks.
However, these works do not take into account the vehicle mobility, which not only affects the channel qualities, but also provides the possibility of frequency sharing among different groups of VUE-pairs \cite{Liu18}.

In this paper, we investigate a Manhattan grid V2V network, where the traffic information changes across the time horizon while the channel quality state depends on the geographical locations of vehicle user equipment (VUE)-transmitter (vTx) and VUE-receiver (vRx) of a VUE-pair.
For such a V2V communication network, the traffic information is time-critical, and hence the fresh traffic updates are of particularly high importance \cite{Abde18, Kaul11}.
An emerging metric for capturing the freshness of information is the age of information (AoI) \cite{Kaul11}.
By definition, AoI is the amount of time elapsed since the most recent information update (at the destination, namely, the vRx of a VUE-pair) was generated (at the source, namely, the corresponding vTx) \cite{Lu18, Kado19}.
It should be noted that optimizing AoI is totally different from queuing delay minimization as in our prior work \cite{Chen19MC}.
The queuing delay can unnecessarily increase the age of an information update \cite{Papp15}.
Motivated by the AoI research and the importance of traffic information freshness, this paper is primarily concerned with the design of an AoI-aware RRM algorithm, which achieves the optimal expected long-term performance for all VUE-pairs.
The following briefly summarizes the main technical contributions of this work.
\begin{itemize}
  \item We formulate the AoI-aware RRM problem in a Manhattan grid V2V network as a single-agent MDP with an infinite horizon discounted criterion, where the roadside unit (RSU) makes decisions regarding frequency band allocation and packet scheduling over time in order to optimize the expected long-term performance for all VUE-pairs.
  \item As the number of VUE-pairs in the network increases, the space of frequency band allocation and packet scheduling decision makings grows exponentially.
      We linearly decompose the MDP, leading to a much simplified decision making procedure.
      The linear decomposition approach allows the VUE-pairs to locally compute the per-VUE-pair Q-functions.
  \item With the consideration of vehicle mobility, the local network state space of all VUE-pairs is extremely huge.
      While the assumption of partial observability at each VUE-pair avoids the overwhelming overhead of information coordination among all VUE-pairs, we propose a proactive algorithm by exploring the recent advances in long short-term memory (LSTM) \cite{Hoch97, Hua19} and DRL \cite{Hass16}, using which VUE-pairs behave optimally with only partial network state observations and without any a priori statistical knowledge of the network dynamics.
      The proposed algorithm includes a centralized offline training at the RSU and a decentralized online testing at the VUE-pairs.
  \item Numerical experiments using TensorFlow \cite{Abad16} are carried out to verify the theoretical studies in this paper, showing that our proposed algorithm outperforms four state-of-the-art baseline algorithms.
\end{itemize}

The remainder of this paper is structured as follows.
In next section, we introduce the considered model of a Manhattan grid V2V communication network and the assumptions used throughout the paper.
In Section \ref{probForm}, we formally formulate the problem of AoI-aware long-term RRM as a single-agent MDP at the RSU and discuss the technical challenges for a general solution.
In Section \ref{probSolv}, we linearly decompose the MDP and propose a proactive algorithm to solve an optimal control policy.
In Section \ref{simuResu}, we present numerical experiments under various parameter settings to compare the performance of the proposed algorithm against other state-of-the-art baseline algorithms.
Finally, we draw the conclusions in Section \ref{conc}.
To ease readability, we list in Table \ref{tabl0} the major notations of this paper.

\begin{table}
\caption{Major notations used in the paper.}\label{tabl0}
\begin{center}
\begin{tabular}{>{\centering}p{2cm}|p{8cm}}
\hline
$K$/$\mathcal{K}$                                     & number/set of VUE-pairs                                               \\\hline
$G$/$\mathcal{G}$                                     & number/set of VUE-pair groups                                         \\\hline
$\mathcal{K}_g$                                       & set of VUE-pairs in group $g$                                         \\\hline
$B$/$\mathcal{B}$                                     & number/set of frequency bands                                         \\\hline
$\mathcal{Y}$                                         & two-dimensional coverage area of RSU                                  \\\hline
$\mathbf{y}_{k, (vTx)}^j$                             & coordinates of vTx of VUE-pair $k$ at slot $j$                        \\\hline
$\mathbf{y}_{k, (vRx)}^j$                             & coordinates of vRx of VUE-pair $k$ at slot $j$                        \\\hline
$\bar{\mathbf{y}}_k^j$                                & coordinates of midpoint of VUE-pair $k$ at slot $j$                   \\\hline
$\ell$                                                & VUE-pair distance                                                     \\\hline
$H_k^j$                                               & channel state of VUE-pair $k$ at slot $j$                             \\\hline
$\mathcal{H}$                                         & channel state space                                                   \\\hline
$\tau$                                                & time duration of one scheduling slot                                  \\\hline
$W$                                                   & bandwidth of a frequency band                                         \\\hline
$\mathbf{f}_k^j$                                      & frequency allocation vector for VUE-pair $k$ at slot $j$              \\\hline
$F_k$, $F_k^j$                                        & frequency allocation indicator for VUE-pair $k$ at slot $j$           \\\hline
$X_k^j$                                               & packet arrivals at VUE-pair $k$ at slot $j$                           \\\hline
$\mathcal{X}$                                         & set of packet arrival states                                          \\\hline
$\lambda$                                             & packet arrival rate                                                   \\\hline
$\mu$                                                 & size of a data packet                                                 \\\hline
$R_k$, $R_k^j$                                        & packet scheduling decision for VUE-pair $k$ at slot $j$               \\\hline
$P_k^j$                                               & transmit power consumption at VUE-pair $k$ at slot $j$                \\\hline
$P_{(max)}$                                           & maximum transmit power                                                \\\hline
$C$, $C_{k, b}^j$                                     & aggregate interference over a frequency band                          \\\hline
$A_k^j$                                               & AoI of VUE-pair $k$ at slot $j$                                       \\\hline
$\mathcal{A}$                                         & set of AoI states                                                     \\\hline
$\mathbf{S}_k$, $\mathbf{S}_k^j$                      & local network state of VUE-pair $k$                                   \\\hline
$\mathbf{S}$, $\mathbf{S}^j$                          & global network state of VUE-pair $k$                                  \\\hline
$\mathcal{S}$                                         & set of local network states                                           \\\hline
$\bm\pi^*$, $\bm\pi$                                  & control policy of RSU                                                 \\\hline
$\pi_{(F)}^*$, $\pi_{(F)}$                            & frequency band allocation policy of RSU                               \\\hline
$\pi_{(R)}^*$, $\pi_{(R)}$                            & packet scheduling policy of RSU                                       \\\hline
$U_k$                                                 & utility function of VUE-pair $k$                                      \\\hline
$U$                                                   & accumulated utility function                                          \\\hline
$\vartheta$                                           & weight of packet drops                                                \\\hline
$\xi$                                                 & weight of AoI                                                         \\\hline
$L_k^j$                                               & packet drops at VUE-pair $k$ at slot $j$                              \\\hline
\end{tabular}
\end{center}
\end{table}

\begin{table*}
\begin{center}
\begin{tabular}{>{\centering}p{2cm}|p{8cm}}
\hline
$V_k$                                                 & expected long-term utility of VUE-pair $k$                            \\\hline
$\gamma$                                              & discount factor                                                       \\\hline
$V$                                                   & state value function of RSU                                           \\\hline
$Q$                                                   & Q-function of RSU                                                     \\\hline
$Q_k$, $\mathds{Q}$                                   & per-VUE-pair Q-function of VUE-pair $k$                               \\\hline
$\mathbf{O}_k$, $\mathbf{O}_k^j$                      & local observation at VUE-pair $k$                                     \\\hline
$\mathcal{O}$                                         & observation set                                                       \\\hline
$\bm\theta$, $\bm\theta^j$, $\bm\theta_{-}^j$         & parameters associated with the DRQN                                   \\\hline
$\mathcal{M}^j$                                       & replay memory at slot $j$                                             \\\hline
$\mathcal{N}^j$                                       & observation pool at slot $j$                                          \\\hline
$\tilde{\mathcal{M}}^j$                               & mini-batch at slot $j$                                                \\
\hline
\end{tabular}
\end{center}
\end{table*}

\section{System Descriptions}
\label{systMode}

In this section, we elaborate on the network and AoI models.
Furthermore, we assume that the RSU clusters the VUE-pairs into multiple groups apart from each other.

\subsection{Network and Channel Models}

\begin{figure}[t]
  \centering
  \includegraphics[width=30pc]{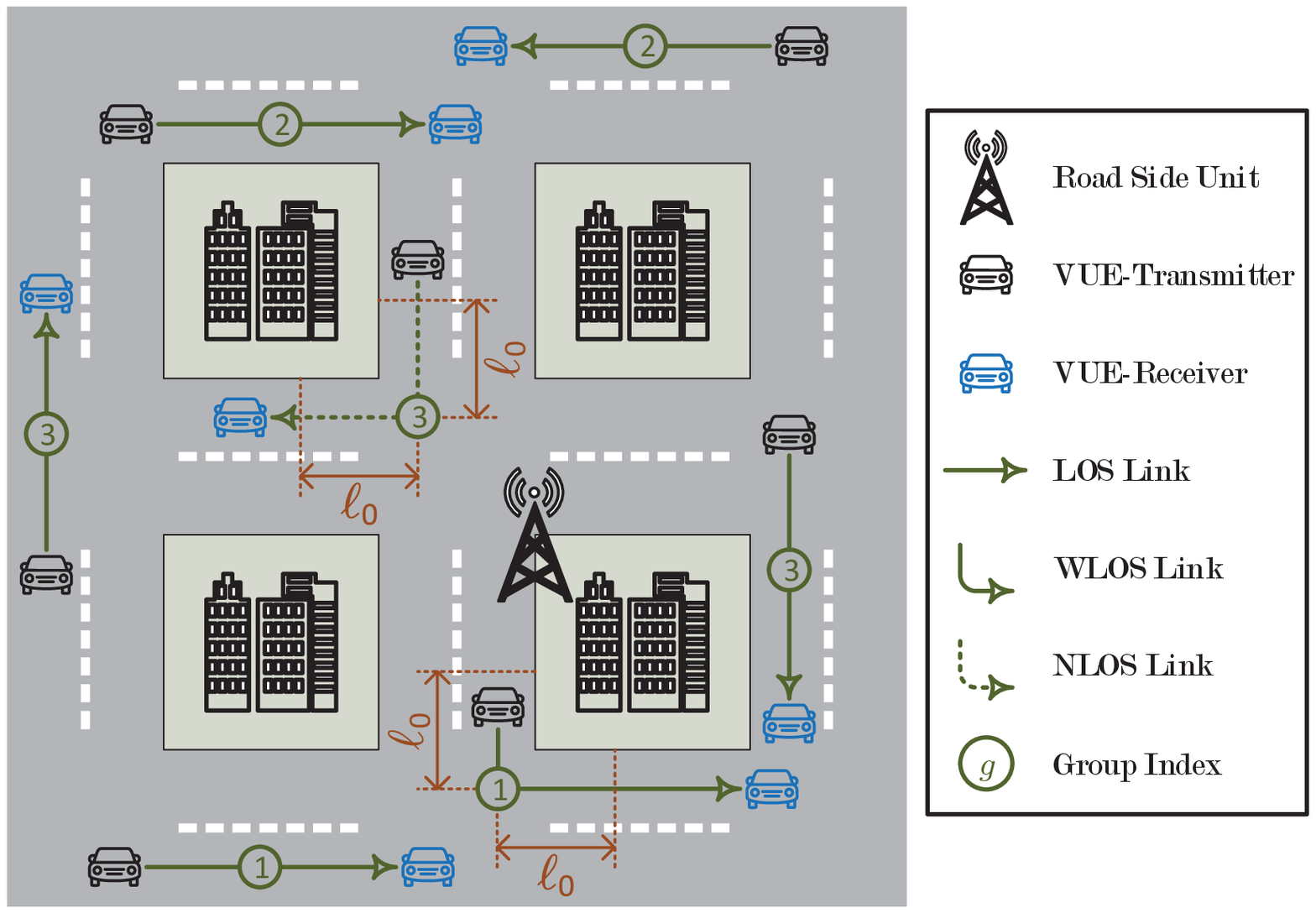}
  \caption{An illustrative Manhattan grid V2V communication network (VUE: vehicle user equipment; LOS: line-of-sight; WLOS: weak-line-of-sight; NLOS: non-line-of-sight.).}
  \label{systModeFigu}
\end{figure}

This work considers the Manhattan grid V2V communications, an illustrative example of which is depicted in Fig. \ref{systModeFigu}.
It has been validated that for a well defined road segment, the vehicle density remains stable \cite{Zhua12}.
Hereinafter, we concentrate on the roads that are covered by a single RSU without loss of generality.
The coverage area can be represented by a two-dimensional Euclidean space $\mathcal{Y}$.
A set $\mathcal{K} = \{1, \cdots, K\}$ of VUE-pairs coexist in the network and share a set $\mathcal{B} = \{1, \cdots, B\}$ of orthogonal frequency bands.
The time horizon is discretized into scheduling slots, with each being of equal time duration $\tau$ and indexed by a positive integer $j \in \mathds{N}_+$.
For each VUE-pair, the vTx always follows at a fixed distance of $\ell$ to the vRx, which moves according to a Manhattan mobility model \cite{Chen18S, Chen19MC}.
We let $\mathbf{y}_{k, (vTx)}^j = \left(y_{k, (vTx)}^{j, 1}, y_{k, (vTx)}^{j, 2}\right)$ and $\mathbf{y}_{k, (vRx)}^j = \left(y_{k, (vRx)}^{j, 1}, y_{k, (vRx)}^{j, 2}\right)$ denote, respectively, the Euclidean coordinates of geographical locations of the vTx and the vRx of a VUE-pair $k \in \mathcal{K}$ within the duration of each scheduling slot $j$.

Depending on whether the vTx and the vRx of a VUE-pair are in the same lane or in the perpendicular lanes, the channel model during each scheduling slot belongs to three categories, namely: 1) line-of-sight (LOS) -- both the vTx and the vRx are in the same lane; 2) weak-line-of-sight (WLOS) -- the vTx and the vRx are in perpendicular lanes and at least one of them is near the intersection within a distance of $\ell_0$; and otherwise, 3) none-line-of-sight (NLOS).
More specifically, the channel state $H_k^j = \psi \cdot h\!\left(\mathbf{y}_{k, (vTx)}^j, \mathbf{y}_{k, (vRx)}^j\right) \in \mathcal{H}$ over the frequency bands experienced by a VUE-pair $k \in \mathcal{K}$ during each slot $j$ includes a fast fading component\footnote{The Rayleigh distribution of the fast fading component is averaged out in this work by applying the previous result in \cite{Chen18S}.} $\psi$ and a path loss $h\!\left(\mathbf{y}_{k, (vTx)}^j, \mathbf{y}_{k, (vRx)}^j\right)$ that applies
\begin{align}\label{pathLossMode}
 h\!\left(\mathbf{y}_{k, (vTx)}^j, \mathbf{y}_{k, (vRx)}^j\right) =
 \left\{
 \begin{array}{l@{~}l}
   \varphi \cdot \left\|\mathbf{y}_{k, (vTx)}^j - \mathbf{y}_{k, (vRx)}^j\right\|_2^{- \eta},               &\mbox{for LOS} \\
   \varphi \cdot \left\|\mathbf{y}_{k, (vTx)}^j - \mathbf{y}_{k, (vRx)}^j\right\|_1^{- \eta},               &\mbox{for WLOS}\\
   \rho    \cdot \left(\left|y_{k, (vTx)}^{j, 1} - y_{k, (vTx)}^{j, 2}\right| \cdot
                   \left|y_{k, (vRx)}^{j, 1} - y_{k, (vRx)}^{j, 2}\right|\right)^{- \eta},                  &\mbox{for NLOS}
 \end{array}
 \right.
\end{align}
for an urban area using 5.9 GHz carrier frequency \cite{Liu18, Mang11}, where we assume that $\mathcal{H}$ is a finite space of the channel states\footnote{The assumption is reasonable due to the discrete-time Manhattan mobility model adopted in this paper.}, $|\cdot|$ is the absolute value norm of a one-dimensional vector, $\|\cdot\|_1$ and $\|\cdot\|_2$ are the $1$-norm and the $2$-norm of a vector, $\eta$ is the path loss exponent, while both $\varphi$ and $\rho$ are the path loss coefficients satisfying $\rho < \varphi \cdot (\ell_0 / 2)^\eta$.

\subsection{AoI Evolution}

vTxs their update time-critical traffic information to the vRxs over the frequency bands on a per-scheduling slot basis.
At the beginning of each scheduling slot, the RSU allocates the limited number of frequency bands to the VUE-pairs in the network.
We assume that during a scheduling slot, a VUE-pair can be assigned at most one frequency band.
Let $\mathbf{f}_k^j = \left(f_{k, b}^j: b \in \mathcal{B}\right)$ denote the frequency allocation vector for each VUE-pair $k \in \mathcal{K}$ at a slot $j$ with
\begin{align}
   f_{k, b}^j
 = \left\{
   \begin{array}{l@{~}l}
      1, & \mbox{if frequency band } b \mbox{ is allocated to VUE-pair } k \mbox{ during scheduling slot } j;         \\
      0, & \mbox{otherwise}.
   \end{array}
   \right.
\end{align}
Let $F_k^j = \sum_{b \in \mathcal{B}} f_{k, b}^j$ be the frequency allocation indicator.
We then have
{
\begin{align}\label{chanCons01}
  F_k^j \in \{0, 1\}, \forall k \in \mathcal{K}.
\end{align}
}
A traffic information update arrives at the vTx of VUE-pair $k$ only at the beginning of each slot.
We assume that the data packets pertaining to the update arrivals are independently distributed over VUE-pairs and identically distributed across scheduling slots with the mean rate of $\lambda$.
For each VUE-pair, the packet arrivals have to be delivered by the end of the scheduling slot and otherwise, the packets will be dropped.
Let $X_k^j$ denote the number of packet arrivals of VUE-pair $k$ at each scheduling slot $j$.
Hence $\textsf{E}\!\left[X_k^j\right] = \lambda$, where the notation $\textsf{E}[\cdot]$ means the expectation of a random variable.

With the frequency allocation $\mathbf{f}_k^j$ during a scheduling slot $j$ by the RSU, the power consumed by the vTx of a VUE-pair $k \in \mathcal{K}$ for the reliable transmission of scheduled $R_k^j$ packets to the corresponding vRx can be computed as
\begin{align}\label{poweCons}
  P_k^j = \sum_{b \in \mathcal{B}} f_{k, b}^j \cdot \dfrac{C_{k, b}^j + W \cdot \sigma^2}{H_k^j} \cdot
          \left(2^{\frac{\mu \cdot R_k^j}{W \cdot \tau}} - 1\right),
\end{align}
where $C_{k, b}^j$ is the received aggregate interference at the vRx of VUE-pair $k$ over a band $b \in \mathcal{B}$ from other VUE-pairs due to frequency sharing, $W$ is the bandwidth of the frequency bands, $\sigma^2$ is the power spectral density of additive white Gaussian background noise, and $\mu$ is the bit size of packet arrivals.
We denote by $P_{(max)}$ the maximum transmit power for all vTxs, that is, $P_k^j \leq P_{(max)}$, $\forall k \in \mathcal{K}$ and $\forall j$.
Thus, $0 \leq R_k^j \leq$ $\min\left\{X_k^j, R_{k, (max)}^j\right\}$, where
\begin{align}
   R_{k, (max)}^j =
   \left\lfloor \left(\tau \cdot \sum_{b \in \mathcal{B}} f_{k, b}^j \cdot W \cdot
   \log_2\left(1 + \frac{H_k^j \cdot P_{(max)}}{C_{k, b}^j + W \cdot \sigma^2}\right)\right) \cdot \frac{1}{\mu}\right\rfloor,
\end{align}
with $\lfloor \cdot \rfloor$ meaning the floor function.
The number of dropped data packets can be tracked by
\begin{align}\label{dropRate}
  L_k^j = X_k^j - F_k^j \cdot R_k^j,
\end{align}
due to the deadline expiry.
The V2V communications rely on fresh data packets.
To this end, the data freshness is quantified by the AoI \cite{Kaul11}.
We designate $A_k^j$ as the AoI of VUE-pair $k$ up to the beginning of scheduling slot $j$, or equivalently, the end of scheduling slot $j - 1$, that is, the time elapsed since the most recently successful packet transmission.
The AoI of VUE-pair $k$ evolves according to
\begin{align}\label{AoIEvol}
  A_k^{j + 1} =
  \left\{
  \begin{array}{l@{~}l}
    \tau,                                      & \mbox{if } F_k^j \cdot R_k^j > 0;                \\
    A_k^j + \tau,                              & \mbox{otherwise}.
  \end{array}
  \right.
\end{align}
Note that by default, we decrease the AoI of a VUE-pair in the network to be $\tau$ at the subsequent scheduling slot if there are any packets delivered during a current slot.

\subsection{VUE-pair Clustering}

To alleviate the mutual interference among the VUE-pairs during packet transmissions and improve the efficiency of frequency band utilization, the RSU clusters the VUE-pairs into a set $\mathcal{G} = \{1, \cdots, G\}$ of disjoint groups based on their geographical locations, where $G > 1$.
The $B$ frequency bands are then exclusively allocated to the VUE-pairs in a same group and reused by the vTxs in different groups.
With a VUE-pair grouping pattern $\left\{\mathcal{K}_g^j: g \in \mathcal{G}\right\}$, the frequency band allocation during each scheduling slot $j$ satisfies an additional constraint, namely,
\begin{align}
  \sum_{k \in \mathcal{K}_g^j} f_{k, b}^j \leq 1, \forall b \in \mathcal{B}, \forall g \in \mathcal{G},        \label{chanCons02}
\end{align}
where $\mathcal{K}_g^j$ denotes the set of VUE-pairs in a group $g \in \mathcal{G}$ during a scheduling slot $j$ and $\cup_{g \in \mathcal{G}} \mathcal{K}_g^j = \mathcal{K}$.
In this work, VUE-pair grouping is done by spectral clustering \cite{Luxb07}.
Denote by $\bar{\mathbf{y}}_k^j = \left(\bar{y}_k^{j, 1}, \bar{y}_k^{j, 2}\right) \triangleq \left(\left(y_{k, (vTx)}^{j, 1} + y_{k, (vRx)}^{j, 1}\right) / 2, \left(y_{k, (vTx)}^{j, 2} + y_{k, (vRx)}^{j, 2}\right) / 2\right)$ the Euclidean coordinates of the midpoint of a VUE-pair $k \in \mathcal{K}$ at scheduling slot $j$.
We choose a distance-based Gaussian similarity matrix $\mathbf{D}^j = \left[d_{k, k'}^j\right]_{k \in \mathcal{K}, k' \in \mathcal{K}}$ to represent the geographical proximity information with each element $d_{k, k'}^j$ being given by
\begin{align}\label{GauSimiEle}
  d_{k, k'}^j =
  \left\{
  \begin{array}{l@{~}l}
    \exp\!\left(- \frac{\left\|\bar{\mathbf{y}}_k^j - \bar{\mathbf{y}}_{k'}^j\right\|_2^2}{\varrho^2}\right),
                        & \mbox{if } \left\|\bar{\mathbf{y}}_k^j - \bar{\mathbf{y}}_{k'}^j\right\|_2 \leq \zeta;   \\
    0,                  & \mbox{otherwise},
  \end{array}
  \right.
\end{align}
where $\zeta$ controls the neighbourhood size and $\varrho$ balances the impact of neighbourhood size.
Accordingly, the VUE-pairs are clustered into groups based on $\mathbf{D}^j$, the procedure of which is described by Algorithm \ref{cluster}.
\begin{algorithm}[t]
    \caption{Spectral Clustering for VUE-pair Grouping at the Beginning of Each Scheduling Slot $j$}
    \label{cluster}
    \begin{algorithmic}[1]
        \STATE Calculate $\mathbf{D}^j$ and a diagonal matrix $\boldsymbol{\Omega}^j = \mathrm{diag}\!\left(\left\{\omega_k^j: k \in \mathcal{K}\right\}\right)$, where the $k$-th diagonal entry of $\boldsymbol{\Omega}^j$ is given by $\omega_k^j = \sum_{k' \in \mathcal{K}} d_{k, k'}^j$.

        \STATE Let $\bm\Phi^j = \left[\bm\phi_g^j: g \in \mathcal{G}\right]$, where $\bm\phi_g^j$ is the eigenvector of the $g$-th smallest eigenvalue of $\mathbf{I} - (\bm\Omega^j)^{- 1/2} \cdot \mathbf{D}^j \cdot (\bm\Omega^j)^{- 1/2}$ with $\mathbf{I}$ being the identity matrix.

        \STATE Use the approach as in \cite{Lloy82} to cluster $K$ normalized row vectors of matrix $\bm\Phi^j$ into $G$ groups.
    \end{algorithmic}
\end{algorithm}
Now we can rewrite the transmit power consumption in (\ref{poweCons}) as
\begin{align}\label{poweConsNew}
  P_k^j = F_k^j \cdot \dfrac{C + W \cdot \sigma^2}{H_k^j} \cdot \left(2^{\frac{\mu \cdot R_k^j}{W \cdot \tau}} - 1\right),
\end{align}
by approximating the received interference from other groups at the vRx of each VUE-pair $k$ over each frequency band $b \in \mathcal{B}$ as a constant $C_{k, b}^j = C$, $\forall j$.

\section{Problem Formulation}
\label{probForm}

In this section, we formulate the problem of AoI-aware RRM in the Manhattan grid V2V network as a discrete-time single-agent MDP with an infinite horizon discounted criterion and discuss the general solution.

\subsection{AoI-Aware RRM}

During each scheduling slot $j$, the local state of a VUE-pair $k \in \mathcal{K}$ can be characterized by $\mathbf{S}_k^j = \left(\left(\mathbf{y}_{k, (vTx)}^j, \mathbf{y}_{k, (vRx)}^j\right), H_k^j, X_k^j, A_k^j\right) \in \mathcal{S} \triangleq \mathcal{Y} \times \mathcal{H} \times \mathcal{X} \times \mathcal{A}$, which includes the information of the geographical location $\left(\mathbf{y}_{k, (vTx)}^j, \mathbf{y}_{k, (vRx)}^j\right)$, the channel state $H_k^j$, the packet arrivals $X_k^j$ and the AoI $A_k^j$.
Herein, $\mathcal{X}$ and $\mathcal{A}$ are the state sets of packet arrivals and AoI.
Let $\mathrm{card}(\cdot)$ denote the cardinality of a set.
We assume that $\mathrm{card}(\mathcal{X})$ and $\mathrm{card}(\mathcal{A})$ are finite, but can be arbitrarily large.
$\mathbf{S}^j = \left(\mathbf{S}_k^j, \mathbf{S}_{- k}^j\right) \in \mathcal{S}^K$ can be used to represent the global network state with $- k$ denoting all the other VUE-pairs in $\mathcal{K}$ without the presence of a VUE-pair $k$.
Let $\bm\pi = \left(\pi_{(F)}, \pi_{(R)}\right)$ be a stationary control policy employed by the RSU, where $\pi_{(F)}$ and $\pi_{(R)}$ are, respectively, the frequency band allocation policy and the packet scheduling policy.
When deploying $\bm\pi$, the RSU observes $\mathbf{S}^j$ at the beginning of a scheduling slot $j$ and accordingly, makes frequency band allocation as well as packet scheduling decisions for the VUE-pairs, that is, $\bm\pi(\mathbf{S}^j) = \left(\pi_{(F)}(\mathbf{S}^j), \pi_{(R)}(\mathbf{S}^j)\right) = (\mathbf{F}^j, \mathbf{R}^j)$, where $\mathbf{F}^j = \left(F_k^j: k \in \mathcal{K}\right)$ and $\mathbf{R}^j = \left(R_k^j: k \in \mathcal{K}\right)$.
We next need an immediate utility function\footnote{To stabilize the offline training of the proactive algorithm developed in Section \ref{probSolv}, the exponential function is chosen to define an immediate utility, whose value does not dramatically diverge.
{Moreover, the exponential function has been well fitted to the generic quantitative relationship between the Quality-of-Experience (QoE) and the Quality-of-Service (QoS) \cite{Fied10}.}} for a VUE-pair $k$ at each slot $j$ as below
\begin{align}\label{utilFunc}
     U_k\!\left(\mathbf{S}^j, F_k^j, R_k^j\right)
   = \exp\!\left(- P_k^j\right) + \vartheta \cdot \exp\!\left(- L_k^j\right)
   + \xi \cdot \exp\!\left(- A_k^j\right),
\end{align}
{ 
to assess the QoE, which is defined as a satisfaction measurement of the QoS parameters including the transmit power consumption, the packet drops as well as the AoI.}
Herein, $\vartheta$ and $\xi$ are the non-negative weighting constants.

From the assumptions on vehicle mobility, packet arrivals as well as AoI evolutions, it can be easily verified that the randomness lying in a sequence of the global network states over the time horizon $\{\mathbf{S}^j: j \in \mathds{N}_+\}$ is Markovian with the following controlled state transition probability \cite{Bell57}
\begin{align}\label{statTranProb}
   \mathbb{P}\!\left(\mathbf{S}^{j + 1} | \mathbf{S}^j, \bm\pi\!\left(\mathbf{S}^j\right)\right) =
 & \prod_{k \in \mathcal{K}} \mathbb{P}\!\left(\left(\mathbf{y}_{k, (vTx)}^{j + 1}, \mathbf{y}_{k, (vRx)}^{j + 1}\right) |
                                               \left(\mathbf{y}_{k, (vTx)}^j, \mathbf{y}_{k, (vRx)}^j\right)\right) \cdot       \nonumber\\
 & \mathbb{P}\!\left(H_k^{j + 1} | \left(\mathbf{y}_k^{j + 1}, \mathbf{y}_k^{j + 1}\right)\right) \cdot
   \mathbb{P}\!\left(X_k^{j + 1}\right) \cdot \mathbb{P}\!\left(A_k^{j + 1} | A_k^j, F_k^j, R_k^j\right),
\end{align}
where $\mathbb{P}(\cdot)$ denotes the probability of an event occurrence.
Given a stationary control policy $\bm\pi$ by the RSU and an initial global network state $\mathbf{S}^1 = \mathbf{S} \in \mathcal{S}^K$, we express the expected long-term discounted utility function $V_k(\mathbf{S}, \bm\pi)$ of each VUE-pair $k \in \mathcal{K}$ as
\begin{align}\label{statValu}
    V_k\!\left(\mathbf{S}, \bm\pi\right) =
    (1 - \gamma) \cdot\textsf{E}_{\bm\pi}\!\!\left[\sum_{j = 1}^\infty (\gamma)^{j  - 1} \cdot
    U_k\!\left(\mathbf{S}^j, F_k^j, R_k^j\right) | \mathbf{S}^1 = \mathbf{S}\right],
\end{align}
where $\gamma \in [0, 1)$ is the discount factor,  $(\gamma)^{k-1}$ denotes the discount factor to the $(k-1)$-th power, and the expectation $\textsf{E}_{\bm\pi}[\cdot]$ is taken over different decision makings under different global network states following a control policy $\bm\pi$ across the scheduling slots.
When $\gamma$ approaches $1$, (\ref{statValu}) approximates the expected long-term un-discounted utility as well \cite{Adel08}, i.e., $\bar{V}_k(\mathbf{S}, \bm\pi) = \textsf{E}_{\bm\pi}\!\! \left[\lim_{J \rightarrow \infty} (1 / J) \cdot \sum_{j = 1}^J U_k\!\left(\mathbf{S}^j, F_k^j, R_k^j\right) | \mathbf{S}^1 = \mathbf{S}\right]$.
We define (\ref{statValu}) as the optimization objective for each VUE-pair $k$ in the network \cite{Paja14}.
Eventually, the AoI-aware RRM problem, which the RSU aims to solve, can be formally formulated as a single-agent MDP, namely, $\forall \mathbf{S} \in \mathcal{S}^K$,
\begin{equation}\label{MDP}
\begin{array}{l}
\begin{array}{l@{~}l}
  \displaystyle\max_{\bm\pi}~V(\mathbf{S}, \bm\pi)\! & = \displaystyle\sum_{k \in \mathcal{K}} V_k\!\left(\mathbf{S}, \bm\pi\right)           \\
                                                     & = (1 - \gamma) \cdot \textsf{E}_{\bm\pi}\!\!\left[\displaystyle\sum_{j = 1}^\infty 
                                                                      (\gamma)^{j  - 1} \cdot
                                                                      U\!\left(\mathbf{S}^j, \bm\pi\!\left(\mathbf{S}^j\right)\right) 
                                                                      \!| \mathbf{S}^1 = \mathbf{S}\right]\!
\end{array} \\
~~\mathrm{s.t.}~~\mbox{constraints (\ref{chanCons01}) and (\ref{chanCons02})},
\end{array}
\end{equation}
where we define $U(\mathbf{S}^j, \bm\pi(\mathbf{S}^j)) = \sum_{k \in \mathcal{K}} U_k\!\left(\mathbf{S}^j, F_k^j, R_k^j\right)$ as the immediate utility from the viewpoint of the RSU, which is accumulated across all VUE-pairs in the network at a scheduling slot $j$.
$V(\mathbf{S}, \bm\pi)$ is also termed as the state value function of a global network state $\mathbf{S}$ under a stationary control policy $\bm\pi$ \cite{Rich98}.

\subsection{General Solution}
\label{optiSolu}

The stochastic optimization problem formulated as in (\ref{MDP}) is a typical infinite-horizon discrete-time single-agent MDP with a discounted criterion.
Denote the optimal stationary control policy by $\bm\pi^* = \left(\pi_{(F)}^*, \pi_{(R)}^*\right)$, which can be obtained through solving the Bellman's equation \cite{Bell03}: $\forall \mathbf{S} \in \mathcal{S}^K$,
\begin{align}\label{BellEqua}
   V(\mathbf{S}) =
   \max_{\bm\pi(\mathbf{S})}\!\left\{(1 - \gamma) \cdot U(\mathbf{S}, \bm\pi(\mathbf{S})) +
   \gamma \cdot \sum_{\mathbf{S}' \in \mathcal{S}^K}  \mathbb{P}(\mathbf{S}' | \mathbf{S},\bm\pi(\mathbf{S})) \cdot V(\mathbf{S}')\right\},
\end{align}
where $V(\mathbf{S}) = V(\mathbf{S}, \bm\pi^*)$ is the optimal state value function and $\mathbf{S}' = \left(\mathbf{S}_k', \mathbf{S}_{- k}'\right) \in \mathcal{S}^K$ is the resulting global network state at a subsequent slot.
A conventional dynamic programming solution to (\ref{BellEqua}) based on the value or policy iteration \cite{Rich98} requires the complete knowledge of network dynamics (\ref{statTranProb}), which is practically challenging.
We then define the right-hand side of (\ref{BellEqua}) as the Q-function
\begin{align}\label{stat_acti_q1}
     Q(\mathbf{S}, \mathbf{F}, \mathbf{R})
   = (1 - \gamma) \cdot U(\mathbf{S}, \mathbf{F}, \mathbf{R})
   + \gamma \cdot \sum_{\mathbf{S}' \in \mathcal{S}^K} \mathbb{P}(\mathbf{S}' | \mathbf{S}, \mathbf{F}, \mathbf{R}) \cdot V(\mathbf{S}'),
\end{align}
where $\mathbf{F} = (F_k: k \in \mathcal{K})$ and $\mathbf{R} = (R_k: k \in \mathcal{K})$ are the decision makings under the current global network state $\mathbf{S}$.
In turn, the optimal state value function $V(\mathbf{s})$ can be straightforwardly obtained from
\begin{align}\label{stat_acti_q2}
    V(\mathbf{S}) = \max_{\mathbf{F}, \mathbf{R}} Q(\mathbf{S}, \mathbf{F}, \mathbf{R}).
\end{align}
By substituting (\ref{stat_acti_q2}) back into (\ref{stat_acti_q1}), we arrive at
\begin{align}
     Q(\mathbf{S}, \mathbf{F}, \mathbf{R})
   = (1 - \gamma) \cdot U(\mathbf{S}, \mathbf{F}, \mathbf{R})
   + \gamma \cdot \sum_{\mathbf{S}' \in \mathcal{S}^K} \mathbb{P}(\mathbf{S}' | \mathbf{S}, \mathbf{F}, \mathbf{R}) \cdot
     \max_{\mathbf{F}', \mathbf{R}'} Q(\mathbf{S}', \mathbf{F}', \mathbf{R}'),
\end{align}
where $\mathbf{F}' = \left(F_k': k \in \mathcal{K}\right)$ and $\mathbf{R}' = \left(R_k': k \in \mathcal{K}\right)$ denote the frequency allocation and packet scheduling decision makings under the subsequent global network state $\mathbf{S}'$.

In a decentralized decision-making procedure as will be later discussed in Section \ref{probSolv}\footnote{In a decentralized decision-making procedure, the VUE-pairs behave with local partial observations of the global network states.}, the Q-learning algorithm \cite{Watk12}, which involves the off-policy updates, is not applicable to finding the optimal per-VUE-pair Q-functions.
Using the state-action-reward-state-action (SARSA) algorithm \cite{Rumm94, Rich98}, the RSU learns the Q-function in an iterative on-policy manner without a priori statistics of the network dynamics.
With observations of the global network state $\mathbf{S} = \mathbf{S}^j$, the decision making $(\mathbf{F}, \mathbf{R}) = (\mathbf{F}^j, \mathbf{R}^j)$, the realized utility $U(\mathbf{S}, \mathbf{F}, \mathbf{R})$ at a current scheduling slot $j$ and the decision making $(\mathbf{F}', \mathbf{R}') = \left(\mathbf{F}^{j + 1}, \mathbf{R}^{j + 1}\right)$ under the global network state $\mathbf{S}' = \mathbf{S}^{j + 1}$ at the next scheduling slot $j + 1$, the RSU updates the Q-function according to the rule
\begin{align}\label{QLearRule}
     Q^{j + 1}(\mathbf{S}, \mathbf{F}, \mathbf{R})
 & = \left(1 - \alpha^j\right) \cdot Q^j(\mathbf{S}, \mathbf{F}, \mathbf{R})                        \nonumber\\
 & + \alpha^j \cdot \left((1 - \gamma) \cdot U(\mathbf{S}, \mathbf{F}, \mathbf{R}) + \gamma \cdot
                        Q^j\!\left(\mathbf{S}', \mathbf{F}', \mathbf{R}'\right)\right),
\end{align}
where $\alpha^j \in [0, 1)$ is the learning rate.
It has been proven that if: i) the global network state transition probability under the optimal stationary control policy $\bm\pi^*$ is time-invariant; ii) $\sum_{j = 1}^\infty \alpha^j$ is infinite and $\sum_{j = 1}^\infty (\alpha^j)^2$ is finite; and iii) all global network state decision making-pairs are visited infinitely often, which can be satisfied by a $\epsilon$-greedy strategy \cite{Rich98}, the SARSA learning process converges and finds $\bm\pi^*$ \cite{Sing00}.
However, two main technical challenges arise as follows:
\begin{enumerate}
  \item under the system model being investigated in this paper, the global network state space $\mathcal{S}^K$ faced by the RSU is \emph{extremely huge}; and
  \item the number $\left((1 + B) \cdot \left(1 + R_{(max)}\right)\right)^K$ of decision makings\footnote{To keep what follows uniform, we do not exclude the infeasible decision makings.} at the RSU grows \emph{exponentially} as $K$ increases, where for the purpose of analysis convenience, we use $R_{(max)}$ as the maximum number of packet departures at the vTxs, i.e., $R_k^j \leq R_{(max)}$, $\forall k \in \mathcal{K}$ and $\forall j$.
\end{enumerate}

\section{A Proactive DRL Algorithm}
\label{probSolv}

\begin{figure}[t]
  \centering
  \includegraphics[width=28pc]{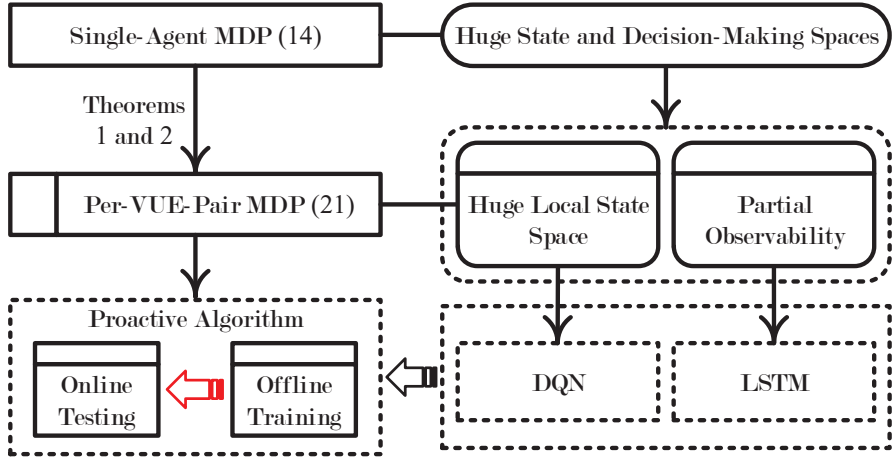}
  \caption{The solution procedure for the AoI-aware RRM problem.}
  \label{solution}
\end{figure}

This section discusses how an optimal control policy can be approached for the RSU.
An overview of our solution is shown in Fig. \ref{solution}.
{
We first decompose the original single-agent MDP as in (\ref{MDP}) into a series of per-VUE-pair MDPs, which can be solved through the decentralized SARSA algorithm.
To address the partial observability and the curse of high dimensionality in local network state space faced by each VUE-pair, we then derive a proactive algorithm based on the LSTM and DRL techniques.
The proactive algorithm is composed of a centralized offline training at the RSU and a decentralized online testing at the VUE-pairs.}

\subsection{Linear Q-function Decomposition}
\label{lineQDeco}

The centralized decision making of frequency band allocation and packet scheduling at a slot from the RSU is executed at the VUE-pairs independently in a decentralized way, based on which we linearly decompose the Q-function (\ref{stat_acti_q1}),
\begin{align}\label{QFuncDeco}
   Q(\mathbf{S}, \mathbf{F}, \mathbf{R}) = \sum_{k \in \mathcal{K}} Q_k(\mathbf{S}, F_k, R_k),
\end{align}
where $Q_k(\mathbf{S}, F_k, R_k)$ is defined to be the per-VUE-pair Q-function for each VUE-pair $k \in \mathcal{K}$ that satisfies
\begin{align}\label{perVUEPQFunc}
     Q_k(\mathbf{S}, F_k, R_k)
 & = (1 - \gamma) \cdot U_k(\mathbf{S}, F_k, R_k)                                                                                   \nonumber\\
 & + \gamma \cdot \sum_{\mathbf{S}' \in \mathcal{S}^K}
     \mathbb{P}\!\left(\mathbf{S}' | \mathbf{S}, \left(F_k, \mathbf{F}_{-k}\right), \left(R_k, \mathbf{R}_{-k}\right)\right) \cdot
     Q_k\!\left(\mathbf{S}', F_k', R_k'\right).
\end{align}
We emphasize that the decision makings across the scheduling slots are performed by each VUE-pair in accordance with the optimal control policy implemented by the RSU.
In other words, $\left(F_k', R_k'\right)$ in (\ref{perVUEPQFunc}), $\forall k \in \mathcal{K}$, under the global network state $\mathbf{S}'$ follows $\bm\pi^*(\mathbf{S}')$, i.e.,
\begin{align}\label{optiActi}
  \bm\pi^*(\mathbf{S}') = \underset{\mathbf{F}'', \mathbf{R}''}{\arg\max} \sum_{k \in \mathcal{K}} Q_k\!\left(\mathbf{S}', F_k'', R_k''\right),
\end{align}
which maximizes the sum of per-VUE-pair Q-functions of all VUE-pairs in the network, where the joint $\mathbf{F}'' = (F_k'': k \in \mathcal{K})$ and $\mathbf{R}'' = (R_k'': k \in \mathcal{K})$ are the possible decision makings under $\mathbf{S}'$.
Particularly, we highlight two key advantages of the linear decomposition approach in the following.
\begin{enumerate}
  \item Simplified decision makings: The linear decomposition motivates the RSU to let the VUE-pairs submit the local per-VUE-pair Q-function values of the frequency band allocation and packet scheduling decisions with the observation of a global network state, based on which the RSU allocates frequency bands and the VUE-pairs then schedule packet transmissions.
      This dramatically reduces the total number of decision makings at the RSU from $\left((1 + B) \cdot \left(1 + R_{(max)}\right)\right)^K$ to $K \cdot \left((1 + B) \cdot \left(1 + R_{(max)}\right)\right)$.
  \item Guaranteed optimal performance: Theorem 1 proves that the linear Q-function decomposition approach guarantees the optimal control policy $\bm\pi^*$.
\end{enumerate}

\emph{Theorem 1:}
The linear Q-function decomposition approach as in (\ref{QFuncDeco}) asserts the optimal expected long-term performance for all VUE-pairs.

\emph{Proof:}
For the Q-function of a centralized decision making $(\mathbf{F}, \mathbf{R})$ under a global network state $\mathbf{S}$ in (\ref{stat_acti_q1}), we have
\begin{align}\label{QDecoOpti}
     Q(\mathbf{S}, \mathbf{F}, \mathbf{R})
 & = (1 - \gamma) \cdot \textsf{E}_{\bm\pi^*}\!\!\!\left[\sum_{j = 1}^\infty (\gamma)^{j - 1} \cdot
     U\!\left(\mathbf{S}^j, \mathbf{F}^j, \mathbf{R}^j\right) |
     \mathbf{S}^1 = \mathbf{S}, \mathbf{F}^1 = \mathbf{F}, \mathbf{R}^1 = \mathbf{R}\right]                                         \nonumber\\
 & = (1 - \gamma) \cdot \textsf{E}_{\bm\pi^*}\!\!\!\left[\sum_{j = 1}^\infty (\gamma)^{j - 1} \cdot
     \sum_{k \in \mathcal{K}} U_k\!\left(\mathbf{S}^j, F_k^j, R_k^j\right) |
     \mathbf{S}^1 = \mathbf{S}, \mathbf{F}^1 = \mathbf{F}, \mathbf{R}^1 = \mathbf{R}\right]                                         \nonumber\\
 & = \sum_{k \in \mathcal{K}} (1 - \gamma) \cdot \textsf{E}_{\bm\pi^*}\!\!\!\left[\sum_{j = 1}^\infty (\gamma)^{j - 1} \cdot
     U_k\!\left(\mathbf{S}^j, F_k^j, R_k^j\right) | \mathbf{S}^1 = \mathbf{S}, F_k^1 = F_k, R_k^1 = R_k\right]                      \nonumber\\
 & = \sum_{k \in \mathcal{K}} Q_k\!\left(\mathbf{S}, F_k, R_k\right),
\end{align}
which completes the proof.
\hfill$\Box$

Therefore, instead of learning the Q-function at the RSU, the SARSA updating rule (\ref{QLearRule}) of the RSU is slightly adapted for each VUE-pair $k \in \mathcal{K}$ to
\begin{align}\label{perVUEQLearRule}
     Q_k^{j + 1}(\mathbf{S}, F_k, R_k)
 & = \left(1 - \alpha^j\right) \cdot Q_k^j\!\left(\mathbf{S}, F_k, R_k\right)           \nonumber\\
 & + \alpha^j \cdot \left((1 - \gamma) \cdot U_k\!\left(\mathbf{S}, F_k, R_k\right) +
     \gamma \cdot Q_k^j\!\left(\mathbf{S}', F_k', R_k'\right)\right).
\end{align}
Theorem 2 ensures the convergence of the decentralized learning process for all VUE-pairs.

\emph{Theorem 2:}
The sequence $\left\{\left(Q_k^j(\mathbf{S}, F_k, R_k): \forall (\mathbf{S}, F_k, R_k), \forall k \in \mathcal{K}\right): j \in \mathds{N}_+\right\}$ by (\ref{perVUEQLearRule}) surely converges to the per-VUE-pair Q-functions $\left(Q_k(\mathbf{S}, F_k, R_k): \forall (\mathbf{S}, F_k, R_k), \forall k \in \mathcal{K}\right)$, if and only if for each VUE-pair $k \in \mathcal{K}$, the $(\mathbf{S}, F_k, R_k)$-pairs are visited for an infinite number of times.

\emph{Proof:}
Since the per-VUE-pair Q-functions are learned simultaneously, we consider the monolithic updates during the decentralized learning process.
That is, the iterative rule in (\ref{perVUEQLearRule}) can be then encapsulated as
\begin{align}\label{SARSA}
     \sum_{k \in \mathcal{K}} Q_k^{j + 1}(\mathbf{S}, F_k, R_k)
 & = \left(1 - \alpha^j\right) \cdot \sum_{k \in \mathcal{K}} Q_k^j\!\left(\mathbf{S}, F_k, R_k\right) \nonumber\\
 & + \alpha^j \cdot \left((1 - \gamma) \cdot \sum_{k \in \mathcal{K}} U_k(\mathbf{S}, F_k, R_k) +
     \gamma \cdot \sum_{k \in \mathcal{K}} Q_k^j(\mathbf{S}', F_k', R_k')\right).
\end{align}
From both sides of (\ref{SARSA}), subtracting the sum of per-VUE-pair Q-functions leads to
\begin{align}\label{SARSAv2}
 & \sum_{k \in \mathcal{K}} Q_k^{j + 1}(\mathbf{S}, F_k, R_k) - \sum_{k \in \mathcal{K}} Q_k(\mathbf{S}, F_k, R_k) =    \nonumber\\
 & \left(1 - \alpha^j\right) \cdot \left(\sum_{k \in \mathcal{K}} Q_k^j(\mathbf{S}, F_k, R_k) -
                                         \sum_{k \in \mathcal{K}} Q_k(\mathbf{S}, F_k, R_k)\right) +
   \alpha^j \cdot T^j(\mathbf{S}, \mathbf{F}, \mathbf{R}),
\end{align}
where
\begin{align}
     T^j(\mathbf{S}, \mathbf{F}, \mathbf{R})
 & = (1 - \gamma) \cdot \sum_{k \in \mathcal{K}} U_k(\mathbf{S}, F_k, R_k)                                                          \nonumber\\
 & + \gamma \cdot \max_{\mathbf{F}'', \mathbf{R}''} \sum_{k \in \mathcal{K}} Q_k^j\!\left(\mathbf{S}', F_k'', R_k''\right) -
     \sum_{k \in \mathcal{K}} Q_k(\mathbf{S}, F_k, R_k))                                                                            \nonumber\\
 & + \gamma \cdot \left(\sum_{k \in \mathcal{K}} Q_k^j(\mathbf{S}', F_k', R_k') -
     \max_{\mathbf{F}'', \mathbf{R}''} \sum_{k \in \mathcal{K}} Q_k^j(\mathbf{S}', F_k'', R_k'')\right).
\end{align}
We let $\Delta^j$, which is given by
\begin{align}\label{learHist}
 & \Delta^j =                                                                                                               \\
 & \left(\left\{\left(\mathbf{S}^\varsigma, \left(F_k^\varsigma, R_k^\varsigma,
                      U_k\!\left(\mathbf{S}^\varsigma, F_k^\varsigma, R_k^\varsigma\right):
                      k \in \mathcal{K}\right)\right): \varsigma \leq j\right\},
                \left(Q_k^j(\mathbf{S}, F_k, R_k): \forall (\mathbf{S}, F_k, R_k), \forall k \in \mathcal{K}\right)\right), \nonumber
\end{align}
denote the history for the first $j$ slots during the decentralized learning process.
The per-VUE-pair Q-functions are $\Delta^j$-measurable, thus both $\left(\sum_{k \in \mathcal{K}} Q_k^{j + 1}(\mathbf{S}, F_k, R_k) - \sum_{k \in \mathcal{K}} Q_k(\mathbf{S}, F_k, R_k)\right)$ and $T^j(\mathbf{S}, \mathbf{F}, \mathbf{R})$ are $\Delta^j$-measurable.
We then attain
\begin{align}\label{sarsaconv}
      & \left\|\textsf{E}\!\!\left[T^j(\mathbf{S}, \mathbf{F}, \mathbf{R}) | \Delta^j\right]\right\|_\infty                 \nonumber\\
 \leq & \left\|\textsf{E}\!\!\left[(1 - \gamma) \cdot \sum_{k \in \mathcal{K}} U_k(\mathbf{S}, F_k, R_k) +
               \gamma \cdot \max_{\mathbf{F}'', \mathbf{R}''} \sum_{k \in \mathcal{K}} Q_k^j(\mathbf{S}', F_k'', R_k'') -
               \sum_{k \in \mathcal{K}} Q_k(\mathbf{S}, F_k, R_k) | \Delta^j\right]\right\|_\infty                          \nonumber\\
    + & \left\|\textsf{E}\!\!\left[\gamma \cdot \left(\sum_{k \in \mathcal{K}} Q_k^j(\mathbf{S}', F_k', R_k') -
               \max_{\mathbf{F}'', \mathbf{R}''} \sum_{k \in \mathcal{K}} Q_k^j(\mathbf{S}', F_k'', R_k'')\right) |
               \Delta^j\right]\right\|_\infty                                                                               \nonumber\\
 \overset{\mbox{(a)}}{\leq}
      & \gamma \cdot \left\|\sum_{k \in \mathcal{K}} Q_k^j(\mathbf{S}, F_k, R_k) -
                            \sum_{k \in \mathcal{K}} Q_k(\mathbf{S}, F_k, R_k)\right\|_\infty                               \nonumber\\
    + & \left\|\textsf{E}\!\!\left[\gamma \cdot \left(\sum_{k \in \mathcal{K}} Q_k^j(\mathbf{S}', F_k', R_k') -
               \max_{\mathbf{F}'', \mathbf{R}''} \sum_{k \in \mathcal{K}} Q_k^j(\mathbf{S}', F_k'', R_k'')\right) |
               \Delta^j\right]\right\|_\infty,
\end{align}
where $\|\cdot\|_\infty$ is the maximum norm of a vector and (a) is due to the convergence property of the standard Q-learning \cite{Watk12}.
We are now left with verifying that $\left\|\textsf{E}\!\left[\gamma \cdot \left(\sum_{k \in \mathcal{K}} Q_k^j(\mathbf{S}', F_k', R_k') -\right.\right.\right.$ $\left.\left.\left.\max_{\mathbf{F}'', \mathbf{R}''} \sum_{k \in \mathcal{K}} Q_k^j(\mathbf{S}', F_k'', R_k'')\right) | \Delta^j\right]\right\|_\infty$ converges to zero, which establishes following: i) an $\epsilon$-greedy policy is deployed to trade-off exploration and exploitation when making frequency band allocation and packet scheduling decisions; ii) the per-VUE-pair Q-function values are upper bounded; and iii) both the global network state and the decision making spaces are finite.
All conditions in \cite[Lemma 1]{Sing00} are met.
Thus the convergence of the decentralized learning is ensured.
\hfill$\Box$

\subsection{Proactive DRL for Optimal Control Policy}
\label{learPoli}

The linear Q-function decomposition developed in previous section brings the immediate advantage of simplified decision makings at the RSU by letting the VUE-pairs to locally learn the per-VUE-pair Q-functions.
However, during the decentralized learning process as in (\ref{perVUEQLearRule}), obtaining a global network state requires exhaustive information coordination between all the VUE-pairs.
{
Note that a specific decision is made by the RSU under a specific global network state.}
Therefore, it is rational for each VUE-pair $k \in \mathcal{K}$ in a stationary network to acquire a partial observation $\mathbf{O}_k^j$ of the private network state $\mathbf{S}_{- k}^j$ at other VUE-pairs during each scheduling slot $j$.
In this work, the partial observation $\mathbf{O}_k^j$ at the beginning of a current scheduling slot $j$ includes the decision making of frequency band allocation as well as packet scheduling from the previous scheduling slot $j - 1$.
That is, $\mathbf{O}_k^j = \left(F_k^{j - 1}, R_k^{j - 1}\right) \in \mathcal{O}$, where $\mathcal{O} = \{0, 1\} \times \left\{0, 1, \cdots, R_{(max)}\right\}$ represents for a VUE-pair the set of partial observations of private network states at other VUE-pairs.
It's worth noting that $\mathrm{card}(\mathcal{O}) \ll \mathrm{card}\!\left(\mathcal{S}^{K - 1}\right)$.

With the local observation $\mathbf{O}_k \in \mathcal{O}$ of $\mathbf{S}_{- k}$ at a current scheduling slot, the per-VUE-pair Q-function (\ref{perVUEPQFunc}) of a VUE-pair $k \in \mathcal{K}$ can be abstracted as
\begin{align}\label{qFuncObse}
  Q_k(\mathbf{S}, F_k, R_k) \approx Q_k(\mathbf{S}_k, \mathbf{O}_k, F_k, R_k), \forall (F_k, R_k),
\end{align}
which does not need to exchange the state information with other VUE-pairs in the network.
Then VUE-pair $k$ learns the Q-function $Q_k(\mathbf{S}_k, \mathbf{O}_k, F_k, R_k)$, $\forall (\mathbf{S}_k, \mathbf{O}_k,F_k, R_k)$, only with local available information.
We can easily find that the local partial observation space $\mathcal{S} \times \mathcal{O}$ is still huge.
The tabular nature in representing Q-function values makes the decentralized SARSA learning in Subsection \ref{lineQDeco} impractical.
Inspired by the widespread success of a deep neural network \cite{Mnih15}, we adopt a double DQN to model the Q-function of a VUE-pair \cite{Hass16}.
Simply using an independent DQN for each VUE-pair raises a new set of technical challenges:
\begin{enumerate}
  \item the possibly asynchronous training of independent DQNs at the VUE-pairs constrains the overall network performance;
  \item in practice, the limited computation capability at a VUE-pair adds another constraint on the feasibility of training a DQN locally; and
  \item the accuracy of (\ref{qFuncObse}), which is based on partial observations, can be, in general, arbitrarily bad.
\end{enumerate}

\begin{figure}[t]
  \centering
  \includegraphics[width=30pc]{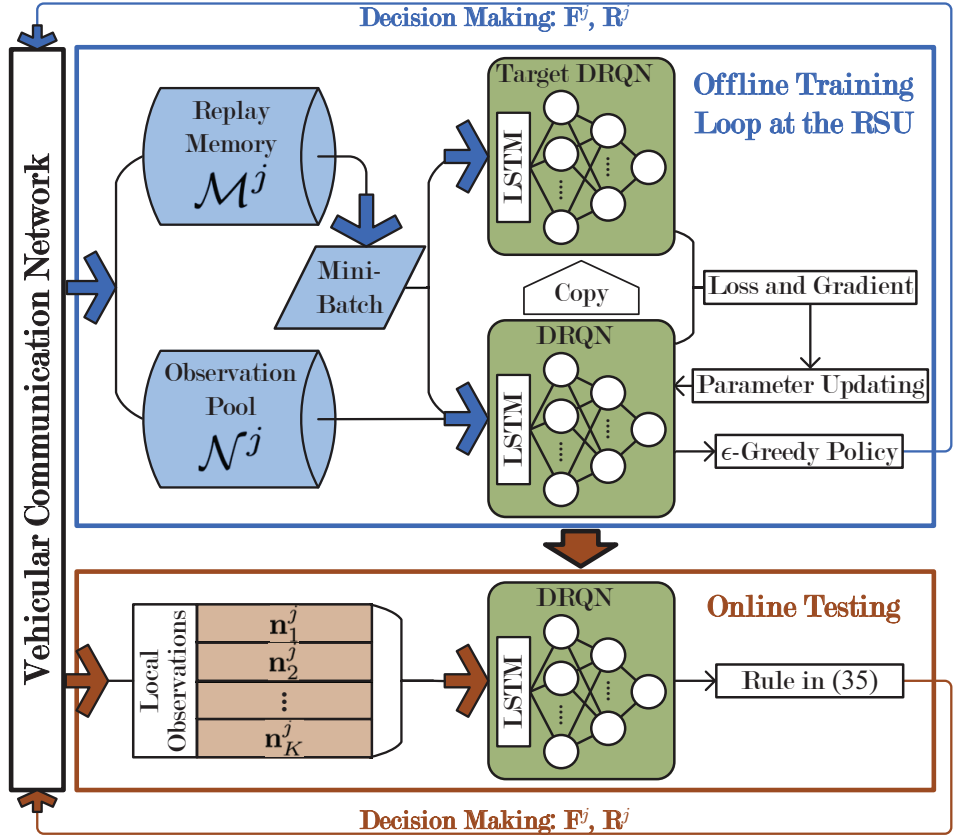}
  \caption{Offline parameter training of the DRQN centralized at the RSU and online control policy testing decentralized at the VUE-pairs.}
  \label{deepLear}
\end{figure}

From the assumptions made throughout this paper and the definition of an identical utility function structure as in (\ref{utilFunc}), there exists homogeneity in the VUE-pairs during the stochastic frequency band allocation and packet scheduling process.
Training a common DQN at the RSU has the potential to be a promising alternative.
Moreover, in order to overcome the partial observability of the global network states, we propose to modify the DQN architecture by leveraging recent advances in recurrent neural networks.
That is, we replace the first fully-connected layer of the DQN with a recurrent LSTM layer \cite{Hoch97, Hua19}, resulting in a deep recurrent Q-network (DRQN) \cite{Hauk15, Napa19}.
Specifically, we let $Q_k(\mathbf{S}, F_k, R_k) \approx \mathds{Q}(\mathbf{n}_k, F_k, R_k; \bm\theta)$, $\forall k \in \mathcal{K}$, where $\mathbf{n}_k = \mathbf{n}_k^j \triangleq \left(\left(\mathbf{S}_k^{j - n + 1}, \mathbf{O}_k^{j - n + 1}\right): n = N, N - 1, \cdots, 1\right)$ consists of the $N$ most recent local partial observations of the global network states up to a current scheduling slot $j$ at VUE-pair $k$ and is taken as the input to the LSTM layer for a more precise and proactive prediction of the current global network state $\mathbf{S}$, while $\bm\theta$ contains a vector of parameters associated with the DRQN.
Considering the computation constraint of the VUE-pairs, the RSU trains the DRQN offline, the procedure of which is illustrated in Fig. \ref{deepLear}.
Instead of finding the per-VUE-pair Q-function, the parameters of the DRQN are learned.

To implement the DRQN offline training, the RSU maintains a replay memory $\mathcal{M}^j$ with the most recent $M$ experiences $\left\{\mathbf{m}^{j - M + 1}, \cdots, \mathbf{m}^j\right\}$ at the beginning of each scheduling slot $j$, where an experience $\mathbf{m}^{j - m + 1}$ ($\forall m \in \{1, \cdots, M\}$) is given as
\begin{align}\label{expe}
     \mathbf{m}^{j - m + 1}
 = & \left(\left(\mathbf{S}_k^{j - m}, \mathbf{O}_k^{j - m}, F_k^{j - m}, R_k^{j - m},
              U_k\!\left(\mathbf{S}^{j - m}, F_k^{j - m}, R_k^{j - m}\right),\right.\right.                                             \nonumber\\
   & \left.\left.~~\mathbf{S}_k^{j - m + 1}, \mathbf{O}_k^{j - m + 1},F_k^{j - m + 1}, R_k^{j - m + 1}\right): k \in \mathcal{K}\right).
\end{align}
Meanwhile, a pool $\mathcal{N}^j = \left\{\mathbf{n}_k^j: k \in \mathcal{K}\right\}$ of $N$ latest partial observations is kept to predict the global network state $\mathbf{S}^j$ for control policy evaluation at slot $j$.
Both $\mathcal{M}^j$ and $\mathcal{N}^j$ from all VUE-pairs in the network are refreshed over the scheduling slots.
The RSU first randomly samples a mini-batch $\tilde{\mathcal{M}}^j = \left\{\breve{\mathcal{M}}^{j_1}, \cdots, \breve{\mathcal{M}}^{j_{\tilde{M}}}\right\}$ of size $\tilde{M}$ from $\mathcal{M}^j$, where each $\breve{\mathcal{M}}^{j_{\tilde{m}}} \nsubseteq \mathcal{M}^j$ ($\forall \tilde{m} \in \left\{1, \cdots, \tilde{M}\right\}$) is given by
\begin{align}\label{batcSamp}
   \breve{\mathcal{M}}^{j_{\tilde{m}}} =
   \left\{\left(\mathbf{n}_k^{j_{\tilde{m}}}, F_k^{j_{\tilde{m}}}, R_k^{j_{\tilde{m}}},
                U_k\!\left(\mathbf{S}^{j_{\tilde{m}}}, F_k^{j_{\tilde{m}}}, R_k^{j_{\tilde{m}}}\right),
                \mathbf{n}_k^{j_{\tilde{m}} + 1}, F_k^{j_{\tilde{m}} + 1}, R_k^{j_{\tilde{m}} + 1}
          \right): k \in \mathcal{K}\right\}.
\end{align}
Then the set $\bm\theta^j$ of parameters at scheduling slot $j$ is updated by minimizing the accumulative loss function, which is defined as
\begin{align}\label{lossFunc}
     \textsf{LOSS}\!\left(\bm\theta^j\right)
 & = \textsf{E}_{\left\{\breve{\mathcal{M}}^{j_{\tilde{m}}} \in \tilde{\mathcal{M}}^j\right\}}\!\!
     \left[\left(\sum_{k \in \mathcal{K}} \left((1 - \gamma) \cdot
     U_k\!\left(\mathbf{S}^{j_{\tilde{m}}}, F_k^{j_{\tilde{m}}}, R_k^{j_{\tilde{m}}}\right)\right.\right.
     \vphantom{\left(\sum_{k \in \mathcal{K}}\right)^2}\right.                                                                              \nonumber\\
 & + \left.\left.\vphantom{\sum_{k \in \mathcal{K}}}\left. \!\! \gamma \cdot 
     \mathds{Q}\!\left(\mathbf{n}_k^{j_{\tilde{m}} + 1}, F_k^{j_{\tilde{m}} + 1}, R_k^{j_{\tilde{m}} + 1}; \bm\theta_{-}^j\right) -
     \mathds{Q}\!\left(\mathbf{n}_k^{j_{\tilde{m}}}, F_k^{j_{\tilde{m}}}, R_k^{j_{\tilde{m}}}; \bm\theta^j\right)\right)\right)^2\right],
\end{align}
where $\bm\theta_{-}^j$ is the set of parameters of the target DRQN at a certain previous scheduling slot before slot $j$.
The gradient is calculated as
\begin{align}\label{grad}
 &   \nabla_{\bm\theta^j} \textsf{LOSS}\!\left(\bm\theta^j\right)                                                                               \nonumber\\
 & = \textsf{E}_{\left\{\breve{\mathcal{M}}^{j_{\tilde{m}}} \in \tilde{\mathcal{M}}^j\right\}}\!\!
     \left[\sum_{k \in \mathcal{K}} \left((1 - \gamma) \cdot
     U_k\!\left(\mathbf{S}^{j_{\tilde{m}}}, F_k^{j_{\tilde{m}}}, R_k^{j_{\tilde{m}}}\right) + \gamma \cdot
     \mathds{Q}\!\left(\mathbf{n}_k^{j_{\tilde{m}} + 1}, F_k^{j_{\tilde{m}} + 1}, R_k^{j_{\tilde{m}} + 1}; \bm\theta_{-}^j\right)\right.\right. \nonumber\\
 & - \left.\left.\mathds{Q}\!\left(\mathbf{n}_k^{j_{\tilde{m}}}, F_k^{j_{\tilde{m}}}, R_k^{j_{\tilde{m}}}; \bm\theta^j\right)\right) \cdot
     \nabla_{\bm\theta^j}\!\! \left(\sum_{k \in \mathcal{K}} \mathds{Q}\!\left(\mathbf{n}_k, F_k, R_k; \bm\theta^j\right)\right)\right].
\end{align}
Algorithm \ref{algo} summarizes the DRQN offline training.
\begin{algorithm}[t]
    \caption{Offline Training of the DRQN at the RSU for RRM in V2V Networks}
    \label{algo}
    \begin{algorithmic}[1]
        \STATE \textbf{initialize} a replay memory $\mathcal{M}^j$ of size $M$, an observation pool $\mathcal{N}^j$ of size $K$, a mini-batch $\tilde{\mathcal{M}}^j$ of size $\tilde{M}$, a decision making $(\mathbf{F}^j, \mathbf{R}^j)$, and a DRQN and a target DRQN with two sets $\bm\theta^j$ and $\bm\theta_{-}^j$ of parameters, for $j = 1$.

        \REPEAT
            \STATE The RSU clusters the $K$ VUE-pairs into $G$ groups based on the geographical locations following Algorithm \ref{cluster} at the beginning of scheduling slot $j$.

            \STATE After performing $(\mathbf{F}^j, \mathbf{R}^j)$ during scheduling slot $j$, each VUE-pair $k \in \mathcal{K}$ realizes a utility $U_k\!\left(\mathbf{S}^j, F_k^j, R_k^j\right)$.

            \STATE Each VUE-pair $k$ observes $\left(\mathbf{S}_k^{j + 1}, \mathbf{O}_k^{j + 1}\right) \in \mathcal{S} \times \mathcal{O}$ at the beginning of next scheduling slot $j + 1$.

            \STATE The RSU updates $\mathcal{N}^{j + 1}$ using $\left\{\left(\mathbf{S}_k^{j + 1}, \mathbf{O}_k^{j + 1}\right): k \in \mathcal{K}\right\}$, which is collected from all VUE-pairs.

            \STATE Under both of the two constraints (\ref{chanCons01}) and (\ref{chanCons02}), the RSU randomly selects $(\mathbf{F}^{j + 1}, \mathbf{R}^{j + 1})$ with a probability of $\epsilon$, or takes the observation pool $\mathcal{N}^{j + 1}$ as the input to the DRQN with $\bm\theta^j$ and determines $(\mathbf{F}^{j + 1}, \mathbf{R}^{j + 1})$ that maximizes the sum $\sum_{k \in \mathcal{K}} \mathds{Q}\!\left(\mathbf{n}_k^{j + 1}, F_k^{j + 1}, R_k^{j + 1}; \bm\theta^j\right)$ with a probability of $1 - \epsilon$.

            \STATE The RSU updates $\mathcal{M}^{j + 1}$ with the most recent experience $\mathbf{m}^{j + 1}$ in the format of (\ref{expe}).

            \STATE With $\tilde{\mathcal{M}}^j$ sampled from $\mathcal{M}^j$, the RSU updates  $\bm\theta^j$ using the gradient given by (\ref{grad}).

            \STATE The RSU regularly resets $\bm\theta_{-}^{j + 1}$ with $\bm\theta^j$.

            \STATE The scheduling slot index is updated by $j \leftarrow j + 1$.
        \UNTIL{A predefined stopping condition is satisfied.}
    \end{algorithmic}
\end{algorithm}

For the online testing, each VUE-pair $k \in \mathcal{K}$ takes $\mathbf{n}_k^j$ at the beginning of each scheduling slot $j$ as an input to the trained DRQN with parameters $\bm\theta$, which outputs the Q-function values $\mathds{Q}\!\left(\mathbf{n}_k^j, F_k, R_k; \bm\theta\right)$, $\forall (F_k, R_k)$.
After receiving $\left(\left(\mathds{Q}\!\left(\mathbf{n}_k^j, F_k, R_k; \bm\theta\right): (F_k, R_k)\right): k \in \mathcal{K}\right)$, the RSU makes the decisions
\begin{align}\label{optiActiDRQN}
  \left(\mathbf{F}^j, \mathbf{R}^j\right) =
  \underset{\mathbf{F}, \mathbf{R}}{\arg\max} \sum_{k \in \mathcal{K}} \mathds{Q}\!\left(\mathbf{n}_k^j, F_k, R_k; \bm\theta\right),
\end{align}
that correspond to the frequency band allocation $\mathbf{F}^j$ and the packet scheduling $\mathbf{R}^j$ for all VUE-pairs while satisfying constraints (\ref{chanCons01}) and (\ref{chanCons02}).

\section{Numerical Experiments}
\label{simuResu}

In this section, we proceed to carry out numerical experiments based on TensorFlow \cite{Abad16} to validate the theoretical studies for the problem of AoI-aware RRM in V2V communications and evaluate the performance achieved from our proposed proactive algorithm.
During the algorithm implementations, the DRQN is first offline trained at the RSU using Algorithm \ref{algo}, after which the online frequency band allocation and packet scheduling decision makings are made according to (\ref{optiActiDRQN}) across the scheduling slots.

\subsection{General Setups}

We use a Manhattan grid model as depicted in Fig. \ref{systModeFigu}, which is with nine intersections in a $250\times250$ m$^2$ area \cite{Chen18S, Liu18}.
In the model, one road consists of two lanes, each of which is in one direction and is of width $4$ m.
The average vehicle speed is $60$ km/h, and the data packets of an update arrival at the vTx of a VUE-pair follow a Poisson arrival process.
For the DRQN, we design two fully connected layers after LSTM layer, each of the three layers containing $32$ neurons\footnote{The tradeoff between the performance improvement and the time spent for the offline training process with a more complex neural network architecture is still an open problem \cite{CCLP18, Chen1802}.}.
ReLU is selected as the activation function \cite{Nair10} and Adam as the optimizer \cite{King15}.
Table \ref{tabl1} lists other parameter values.
\begin{table}
  \caption{Parameter values in experiments.}\label{tabl1}
        \begin{center}
        \begin{tabular}{|c|c||c|c|}
              \hline
              Parameter     & Value                 & Parameter     & Value                             \\\hline
              \hline
              $\varphi$     & $-68.5$ dB            & $\rho$        & $-54.5$ dB                        \\\hline
              $\eta$        & $1.61$                & $\ell_0$      & $15$ m                            \\\hline
              $\zeta$       & $150$ m               & $\varrho$     & $30$ m                            \\\hline
              $G$           & $10$                  & $W$           & $800$ kHz                         \\\hline
              $\sigma^2$    & $-174$ dBm/Hz         & $\tau$        & $3$ ms                            \\\hline
              $\vartheta$   & $2$                   & $\xi$         & $0.9$                             \\\hline
              $\gamma$      & $0.9$                 & $\mu$         & $2$ kb                            \\\hline
              $P_{(max)}$   & $2$ W                 & $N$           & $10$                              \\\hline
              $M$           & $5000$                & $\breve{M}$   & $200$                             \\
              \hline
        \end{tabular}
    \end{center}
\end{table}

For the performance comparisons, we consider the following four baseline algorithms as well.
\begin{enumerate}
  \item \emph{Channel-Aware:} At the beginning of a scheduling slot, the RSU allocates the frequency bands to the groups of VUE-pairs based on the channel state information.
  \item \emph{Packet-Aware:} According to the packet arrival status at the beginning of a slot, the RSU allocates the frequency bands to the VUE-pairs of different groups.
  \item \emph{AoI-Aware:} The RSU allocates the frequency bands at each scheduling slot according to the AoI of VUE-pairs from different groups.
  \item \emph{Random:} With this baseline, the RSU first randomly allocates the frequency bands to the groups of VUE-pairs, and then schedules a random number of packets for the vTx of each VUE-pair.
\end{enumerate}
During the implementation of first three baselines, after the centralized frequency band allocation at the RSU at the beginning of each scheduling slot $j$, the vTx of each VUE-pair $k \in \mathcal{K}$ transmits a maximum feasible number $\min\!\left\{X_k^j, R_{k, (max)}^j\right\}$ of packets \cite{Lu18}.

\subsection{Experiment Results}
\subsubsection{Efficiency of Offline Training}

\begin{figure}[t]
  \centering
  \includegraphics[width=0.48\textwidth]{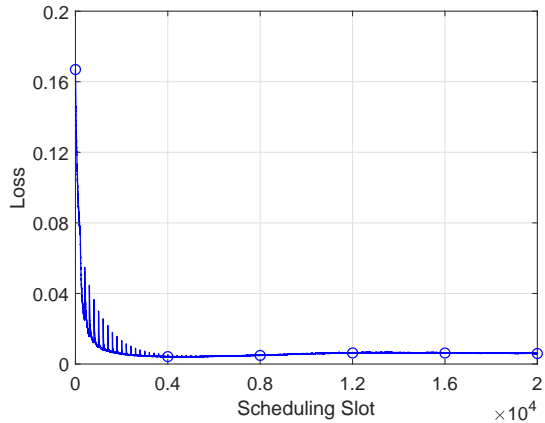}
  \caption{Illustration of the convergence speed of the offline training: $K = 56$, $B = 5$, $\ell = 50$ m and $\lambda = 5$ packets/slot.}
  \label{sim01}
\end{figure}

The aim of first experiment is to examine the offline training efficiency of a common DRQN centralized at the RSU, for which the metric can be the convergence speed of the loss function (\ref{lossFunc}) during the training procedure as in Algorithm \ref{algo}.
In the experiment, we select the number of VUE-pairs and the packet arrival rate as $K = 56$ and $\lambda = 5$ packets/slot, respectively.
The distance between the vTx and the vRx of each VUE-pair is fixed to be $\ell = 50$ m and the number of frequency bands is chosen as $B = 5$.
We plot the variations in $\textsf{LOSS}(\theta^j)$ across the scheduling slots in Fig. \ref{sim01}, which tells that the offline training procedure converges within $1.2 \cdot 10^4$ scheduling slots.
The slow convergence speed also confirms that the offline training is costly and needs to be done from the RSU side.

\subsubsection{Performance with Changing Number of Bands}
\label{simu02}

\begin{figure}[!htb]\centering
   \begin{minipage}{0.48\textwidth}
     \centering
     \includegraphics[width=\textwidth]{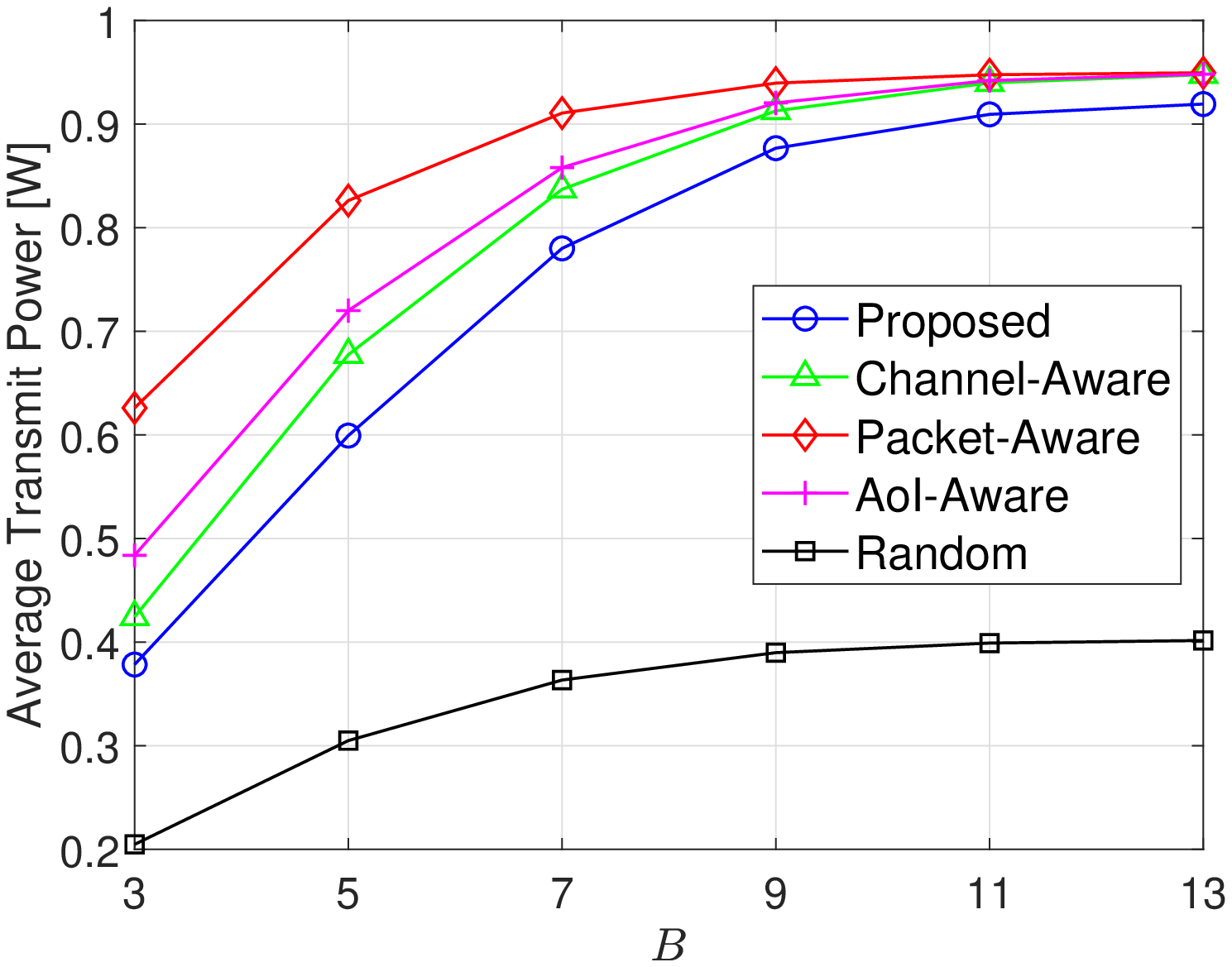}
     \caption{Average transmit power per VUE-pair across the time horizon versus $B$: $K = 56$, $\ell = 50$ m and $\lambda = 5$ packets/slot.}
     \label{sim02_01}
   \end{minipage}
   \hspace{0.1cm}
   \begin{minipage}{0.48\textwidth}
     \centering
     \includegraphics[width=\textwidth]{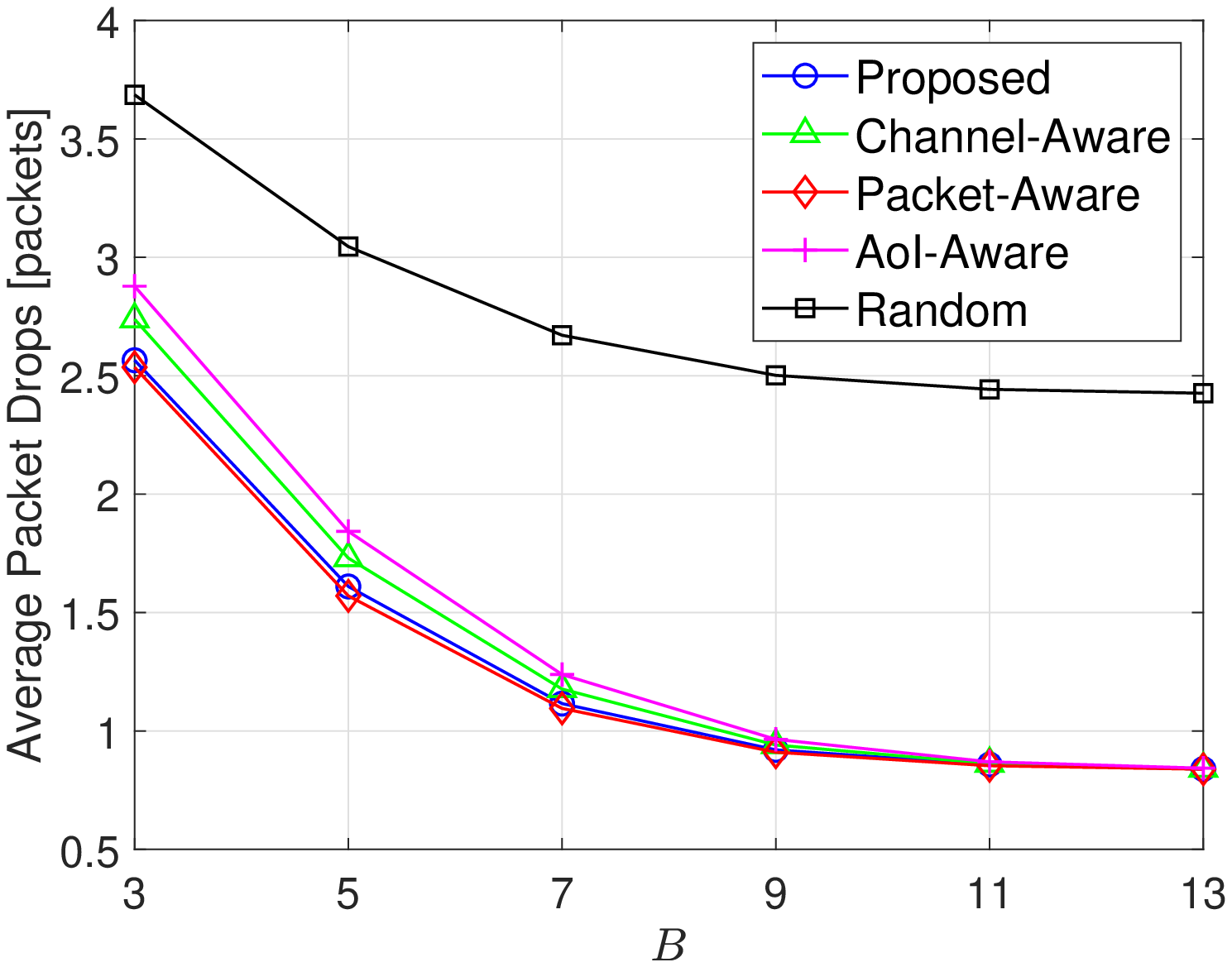}
     \caption{Average packet drops per VUE-pair across the time horizon versus $B$: $K = 56$, $\ell = 50$ m and $\lambda = 5$ packets/slot.}
     \label{sim02_02}
   \end{minipage}
   \begin{minipage}{0.48\textwidth}
     \centering
     \vspace{1cm}
     \includegraphics[width=\textwidth]{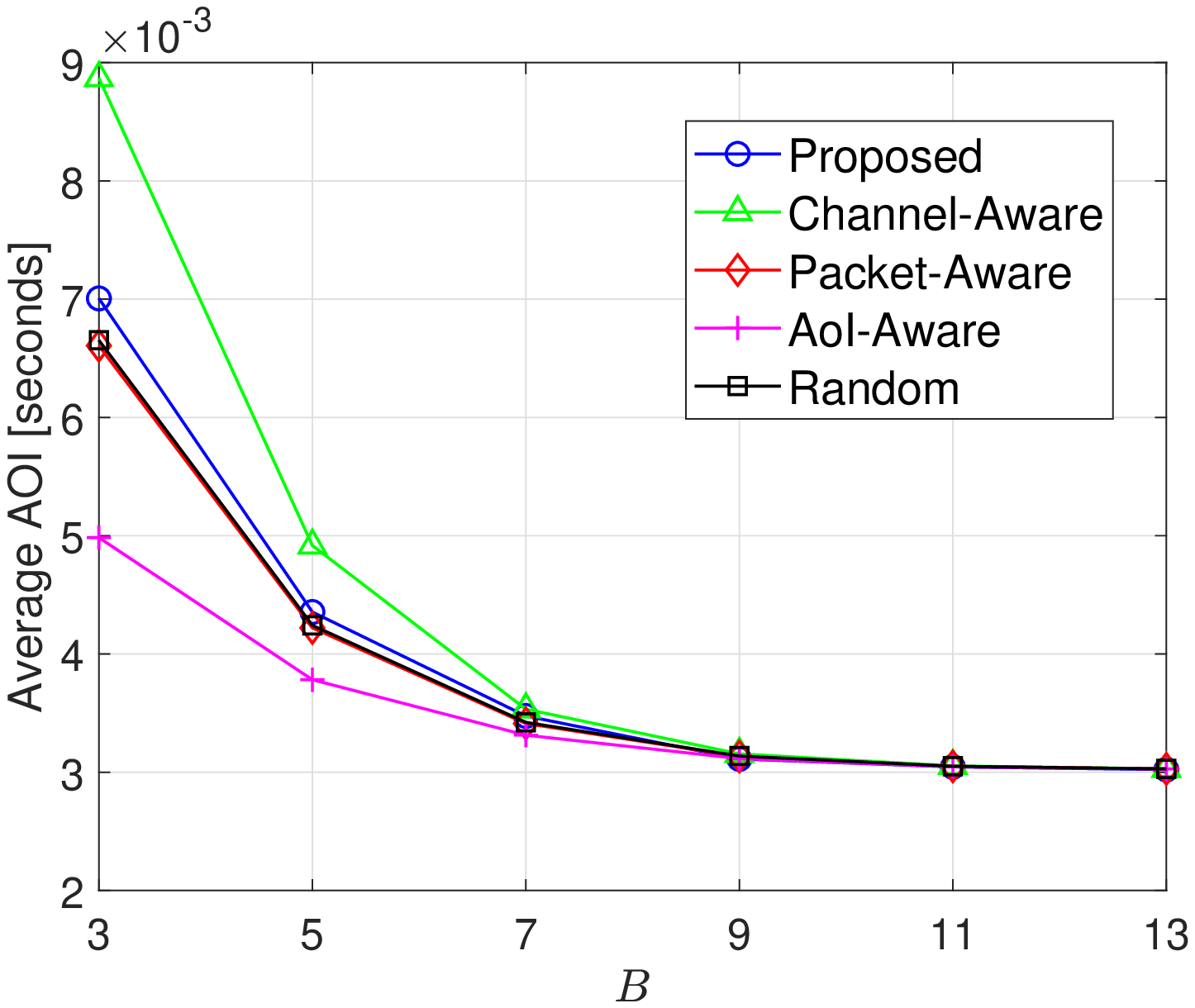}
     \caption{Average AoI per VUE-pair across the time horizon versus $B$: $K = 56$, $\ell = 50$ m and $\lambda = 5$ packets/slot.}
     \label{sim02_03}
   \end{minipage}
   \hspace{0.1cm}
   \begin{minipage}{0.48\textwidth}
     \centering
     \vspace{1cm}
     \includegraphics[width=\textwidth]{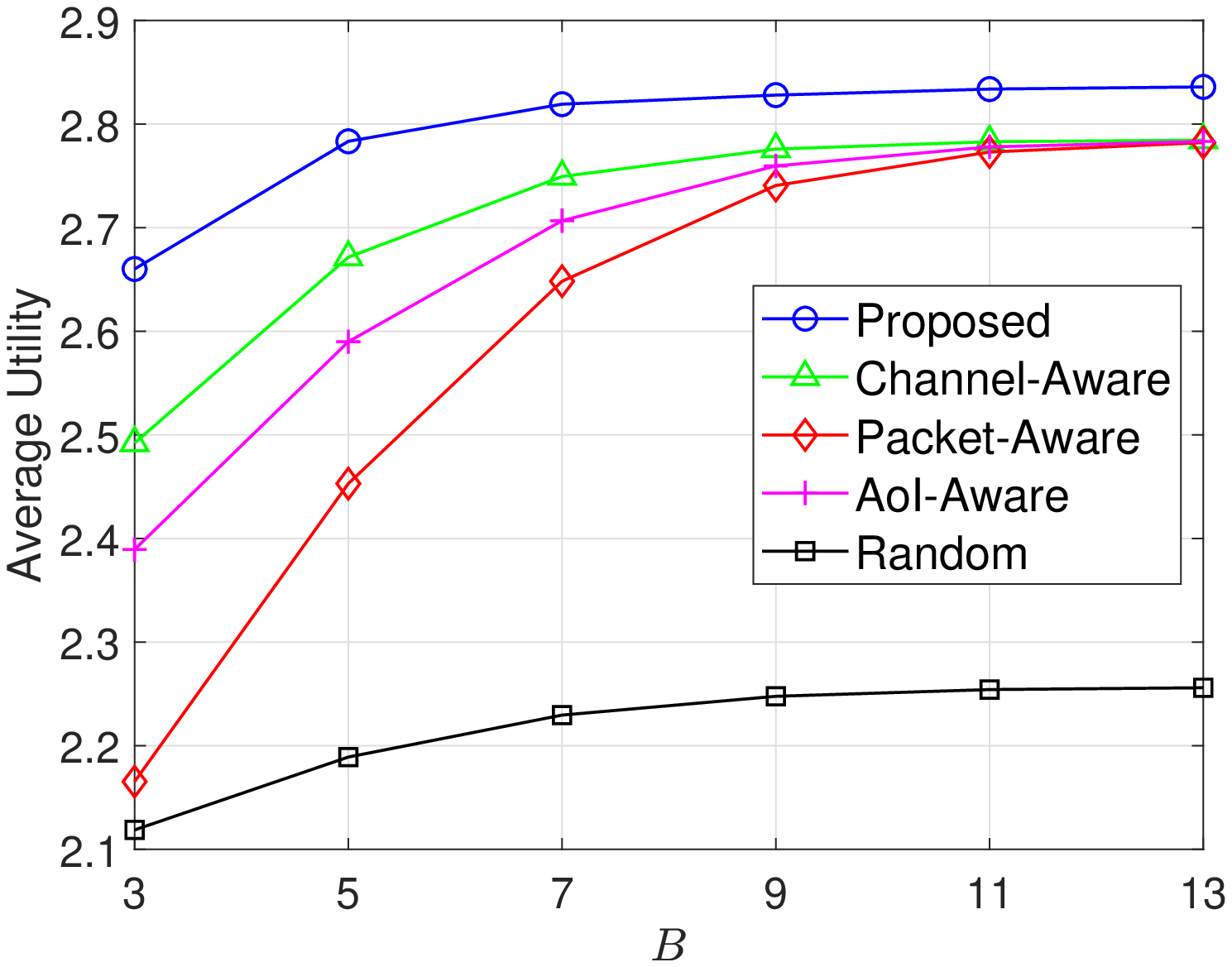}
     \caption{Average utility per VUE-pair across the time horizon versus $B$: $K = 56$, $\ell = 50$ m and $\lambda = 5$ packets/slot.}
     \label{sim02_04}
   \end{minipage}
\end{figure}





Next, we demonstrate the average performance per scheduling slot in terms of the average transmit power consumption, the average packet drops, the average AoI and the average utility by changing the number $B$ of frequency bands.
{The average performance per VUE-pair across the time horizon has been commonly used, for example, in works \cite{Liu18, Chen19M}}.
Other parameters values are the same as in the previous experiment.
The results are depicted in Figs. \ref{sim02_01}, \ref{sim02_02}, \ref{sim02_03} and \ref{sim02_04}.
Fig. \ref{sim02_01} illustrates the average transmit power consumption per VUE-pair per scheduling slot, Fig. \ref{sim02_02} illustrates the average packet drops per VUE-pair per scheduling slot, Fig. \ref{sim02_03} illustrates the AoI per VUE-pair per scheduling slot, while Fig. \ref{sim02_04} illustrates the average utility per VUE-pair per scheduling slot.

In each plot, we compare the performance from the proposed proactive algorithm with other four baseline algorithms.
It is obvious that more frequency bands suggest more opportunities for the VUE-pairs to deliver the arriving data packets, which explains that the average transmit power consumption increases as the number of frequency bands increases (as shown in Fig. \ref{sim02_01}).
At the same time, both the average number of packet drops (as shown in Fig. \ref{sim02_02}) and the average AoI (as shown in Fig. \ref{sim02_03}) decrease.
Given the weight values as in Table \ref{tabl1}, the packet drops and the AoI jointly dominate the utility function (\ref{utilFunc}).
It can be hence observed from Fig. \ref{sim02_04} that as the number of frequency bands increases, the average utility performance improves for all algorithms.
{
When the value of $B$ increases to be large enough such that the vTx of a VUE-pair can always be allocated one frequency band for updating the fresh data packets to the corresponding vRx, the average performance saturates because of the maximum transmit power constraint while in particular, all curves in Fig. \ref{sim02_03} converge to a common value.}
More importantly, the average utility performance from the proposed algorithm outperforms all other baselines, indicating that the proposed algorithm realizes a better trade-off between the transmit power consumption, the packet drops and the AoI.

\subsubsection{Performance under Various VUE-Pair Distance}
\label{simu03}

\begin{figure}[!htb]\centering
   \begin{minipage}{0.48\textwidth}
     \centering
     \includegraphics[width=\textwidth]{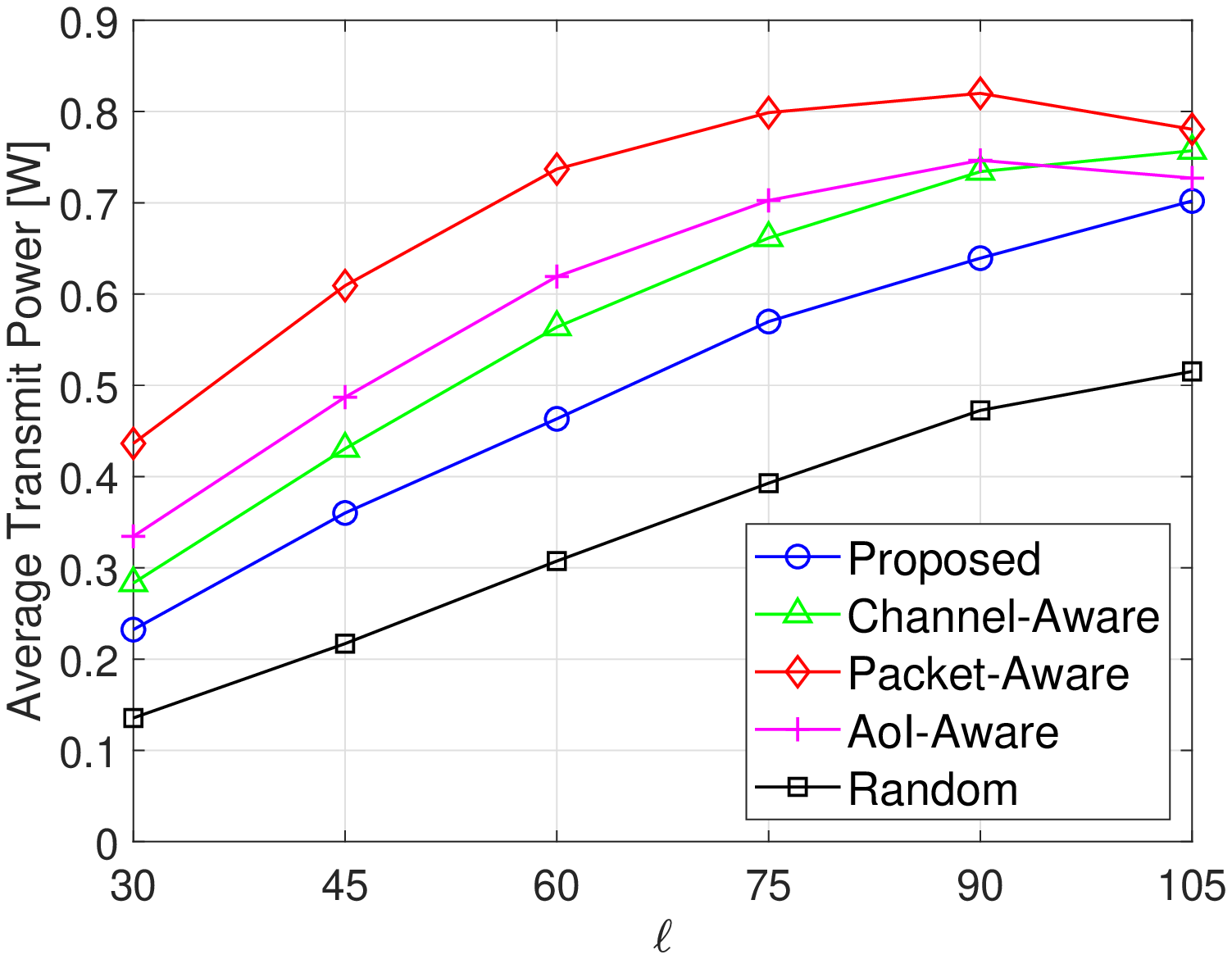}
     \caption{Average transmit power per VUE-pair across the time horizon versus $\ell$ m: $K = 68$, $B =5$ and $\lambda = 4$ packets/slot.}
     \label{sim03_01}
   \end{minipage}
   \hspace{0.1cm}
   \begin{minipage}{0.48\textwidth}
     \centering
     \includegraphics[width=\textwidth]{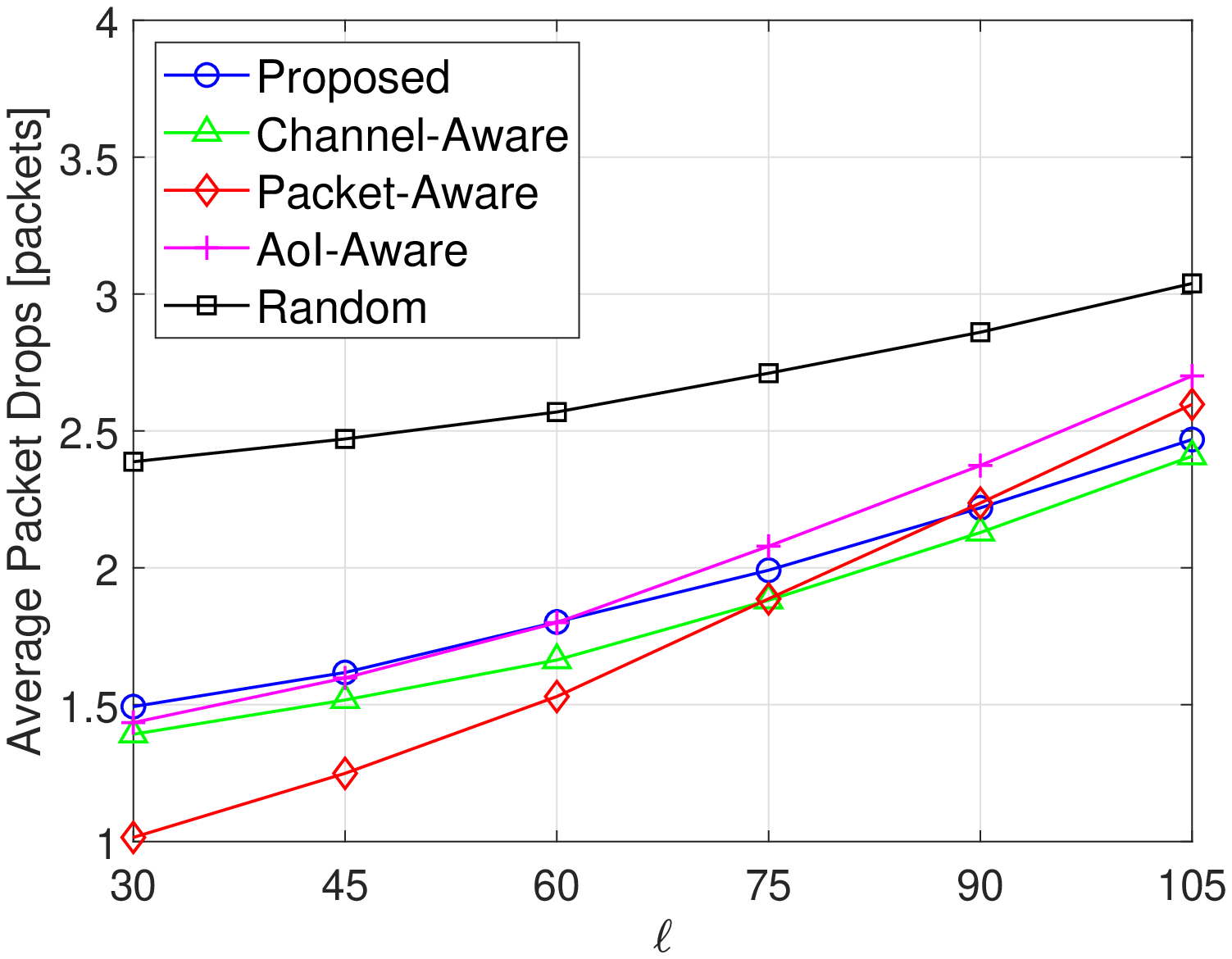}
     \caption{Average packet drops per VUE-pair across the time horizon versus $\ell$ m: $K = 68$, $B =5$ and $\lambda = 4$ packets/slot.}
     \label{sim03_02}
   \end{minipage}
   \begin{minipage}{0.48\textwidth}
     \centering
     \vspace{1cm}
     \includegraphics[width=\textwidth]{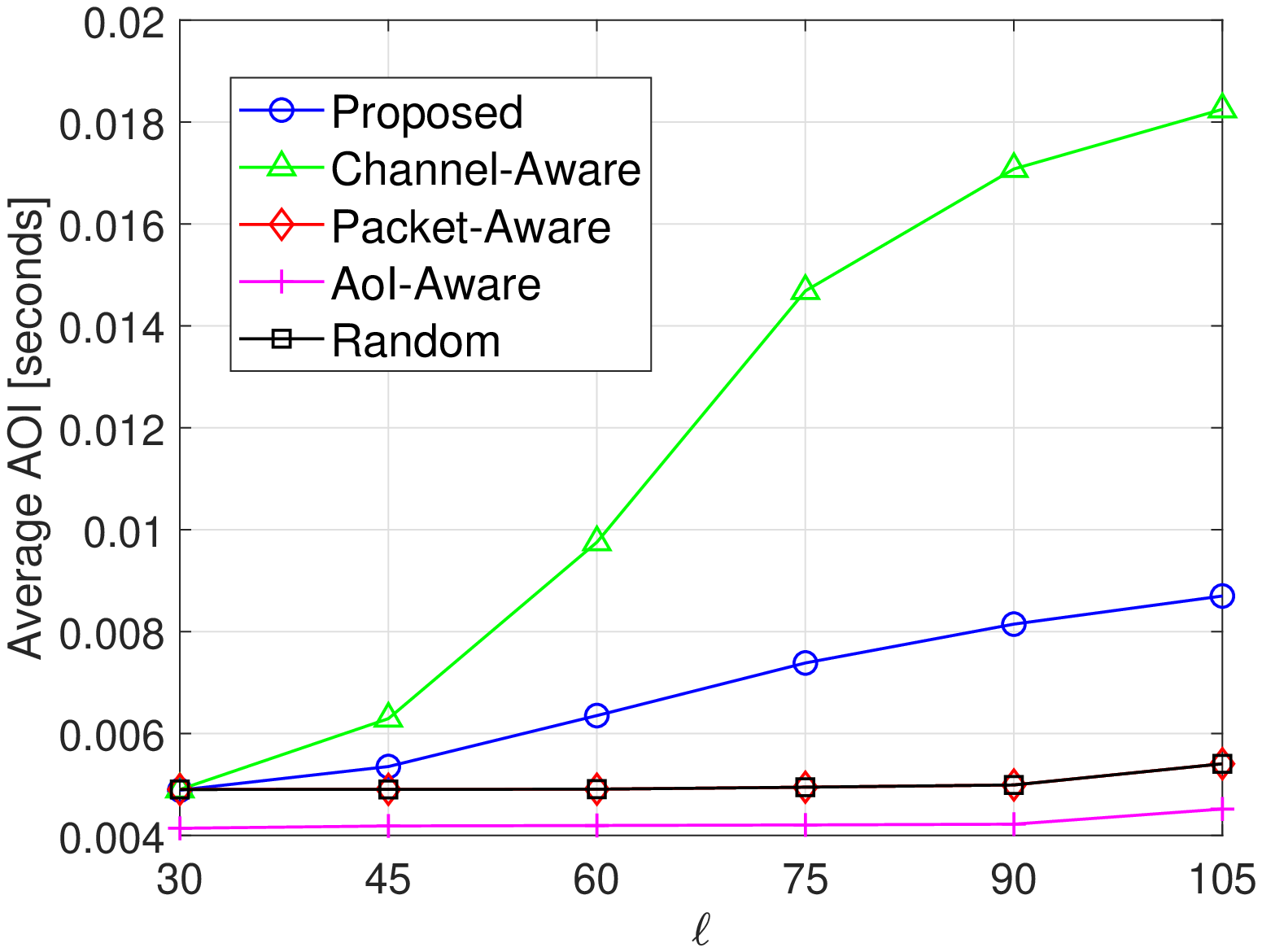}
     \caption{Average AoI per VUE-pair across the time horizon versus $\ell$ m: $K = 68$, $B =5$ and $\lambda = 4$ packets/slot.}
     \label{sim03_03}
   \end{minipage}
   \hspace{0.1cm}
   \begin{minipage}{0.48\textwidth}
     \centering
     \vspace{1cm}
     \includegraphics[width=\textwidth]{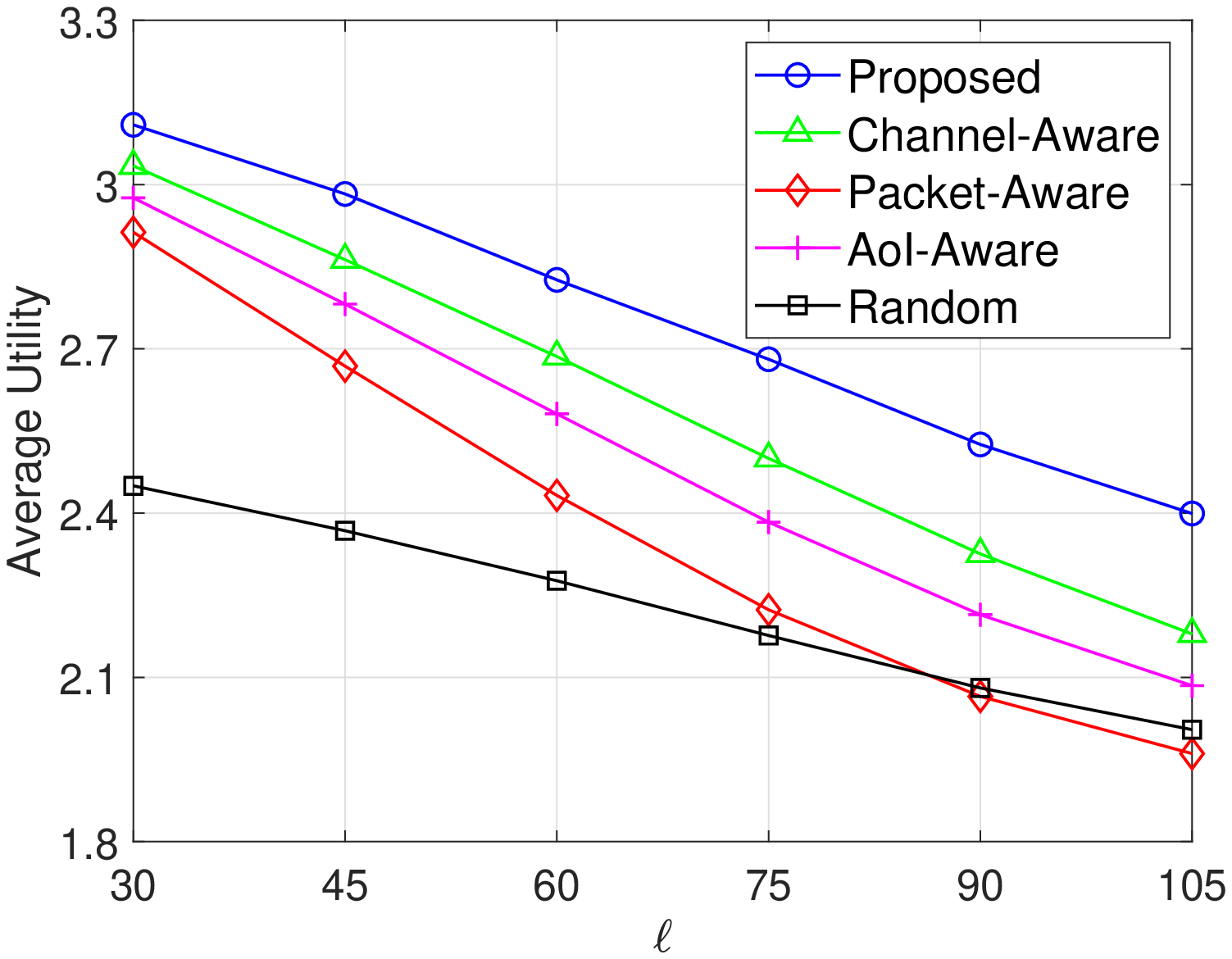}
     \caption{Average utility per VUE-pair across the time horizon versus $\ell$ m: $K = 68$, $B =5$ and $\lambda = 4$ packets/slot.}
     \label{sim03_04}
   \end{minipage}
\end{figure}





We then exhibit the average transmit power consumption, the average packet drops, the average AoI and the average utility per VUE-pair under different values of $\ell$ in Figs. \ref{sim03_01}, \ref{sim03_02}, \ref{sim03_03} and \ref{sim03_04}.
We configure the parameters in this experiment as: $K = 68$, $B = 5$ and $\lambda = 4$ packets/slot.
We can see from the curves in Fig. \ref{sim03_04} that as the VUE-pair distance increases, the VUE-pairs receive a smaller utility on average.
When the distance between the vTx and the vRx of a VUE-pair increases, the channel quality becomes worse.
Thus transmitting the same number of packets requires more power consumption (as shown in Fig. \ref{sim03_01}) and limited by the transmit power at the vTxs, the frequency bands become unavailable for data transmissions more frequently across scheduling slots, resulting in increased average packet drops (as shown in Fig. \ref{sim03_02}) and average AoI (as shown in Fig. \ref{sim03_03}).
Since the Packet-Aware and the AoI-Aware baseline algorithms do not account for the channel qualities, the increase of VUE-pair distance increases the possibility of allocating a frequency band to a VUE-pair with a bad channel state, {under which it is impossible to deliver data packets even with the maximum transmit power}.
This is the reason why the realized average transmit power consumption of both algorithms first increases and then decreases.

\subsubsection{Performance of Different Number of VUE-Pairs}

\begin{figure}[!htb]\centering
   \begin{minipage}{0.48\textwidth}
     \centering
     \includegraphics[width=\textwidth]{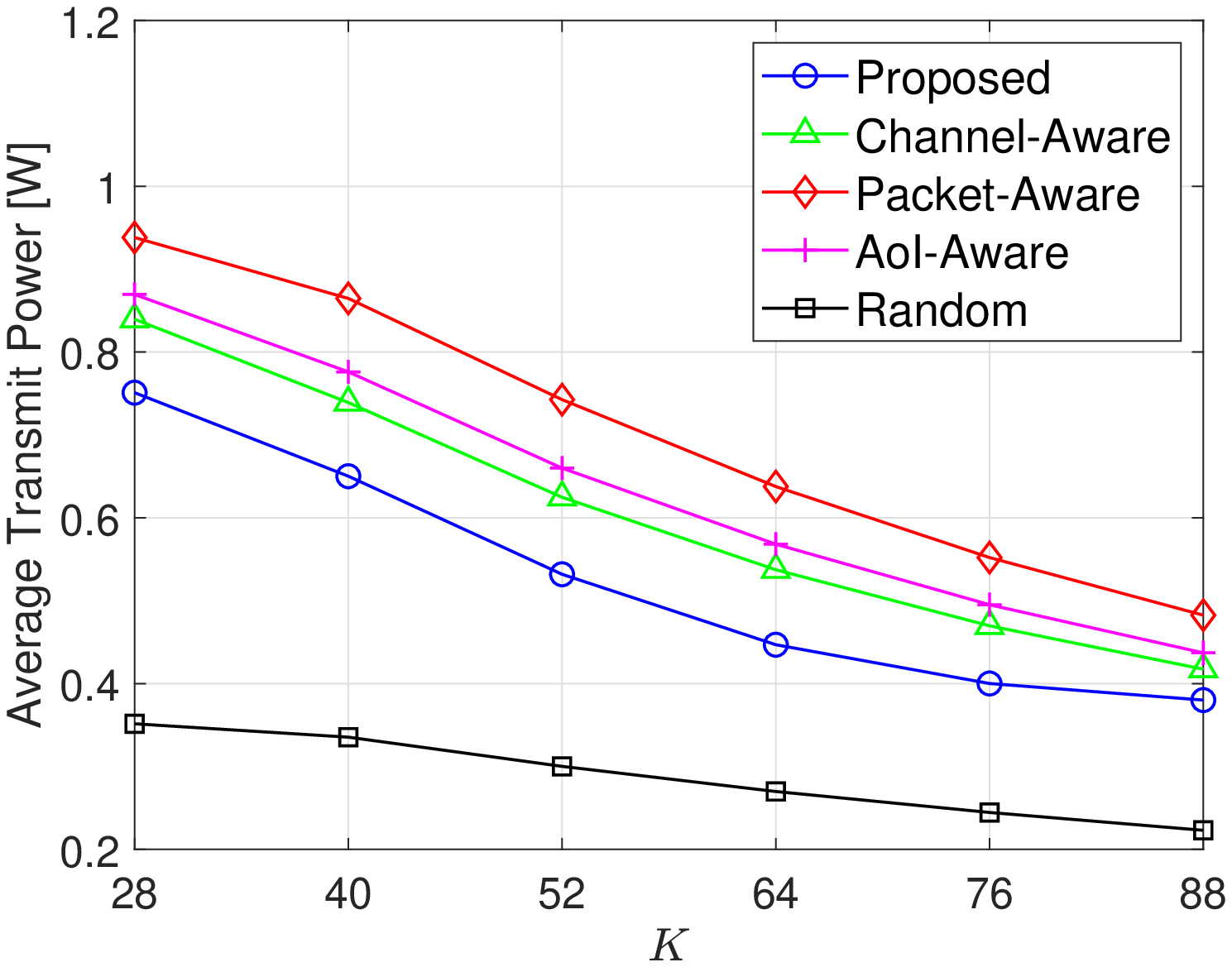}
     \caption{Average transmit power per VUE-pair across the time horizon versus $K$: $B = 3$, $\ell = 65$ m and $\lambda = 6$ packets/slot.}
     \label{sim04_01}
   \end{minipage}
   \hspace{0.1cm}
   \begin{minipage}{0.48\textwidth}
     \centering
     \includegraphics[width=\textwidth]{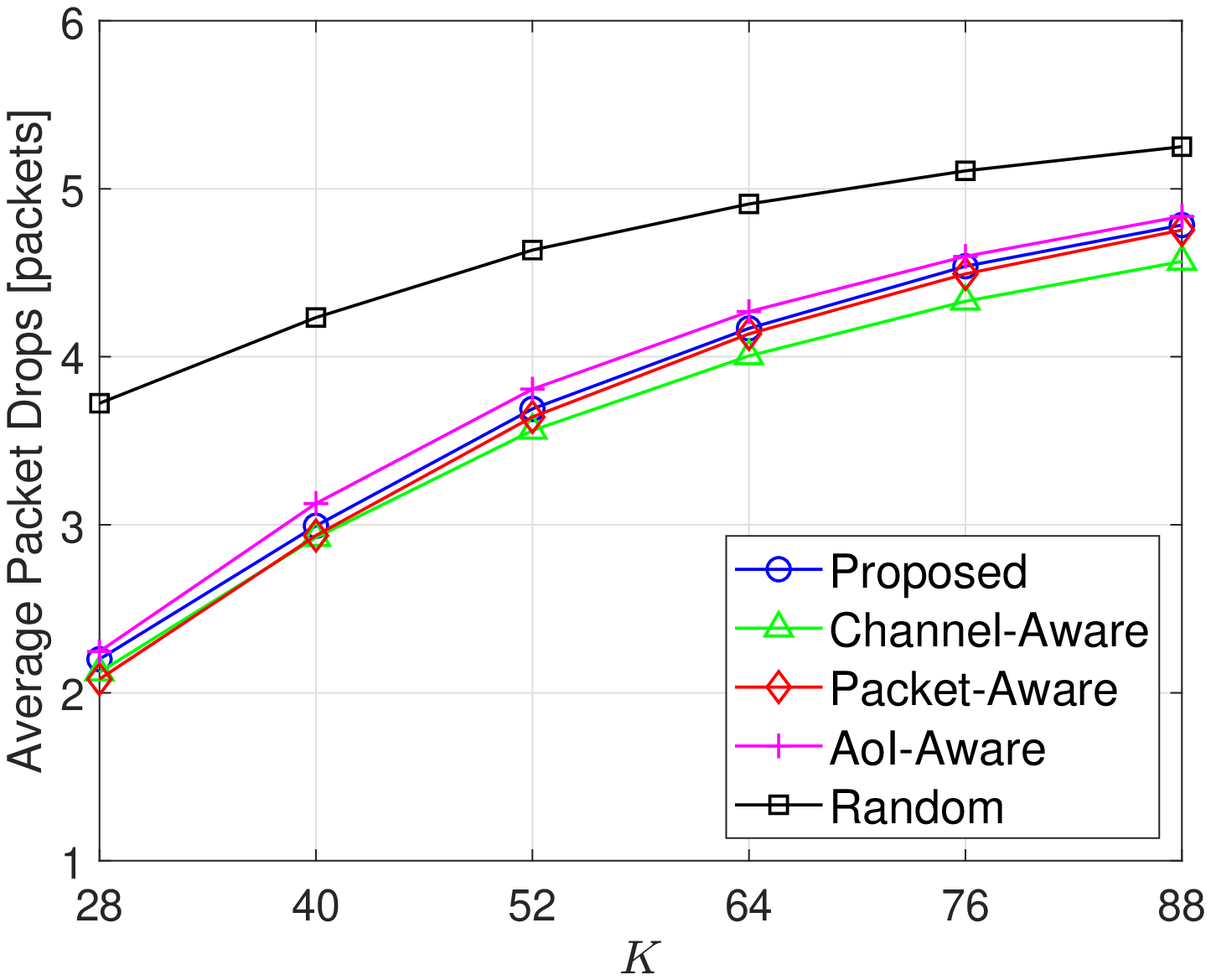}
     \caption{Average packet drops per VUE-pair across the time horizon versus $K$: $B = 3$, $\ell = 65$ m and $\lambda = 6$ packets/slot.}
     \label{sim04_02}
   \end{minipage}
   \begin{minipage}{0.48\textwidth}
     \centering
     \vspace{1cm}
     \includegraphics[width=\textwidth]{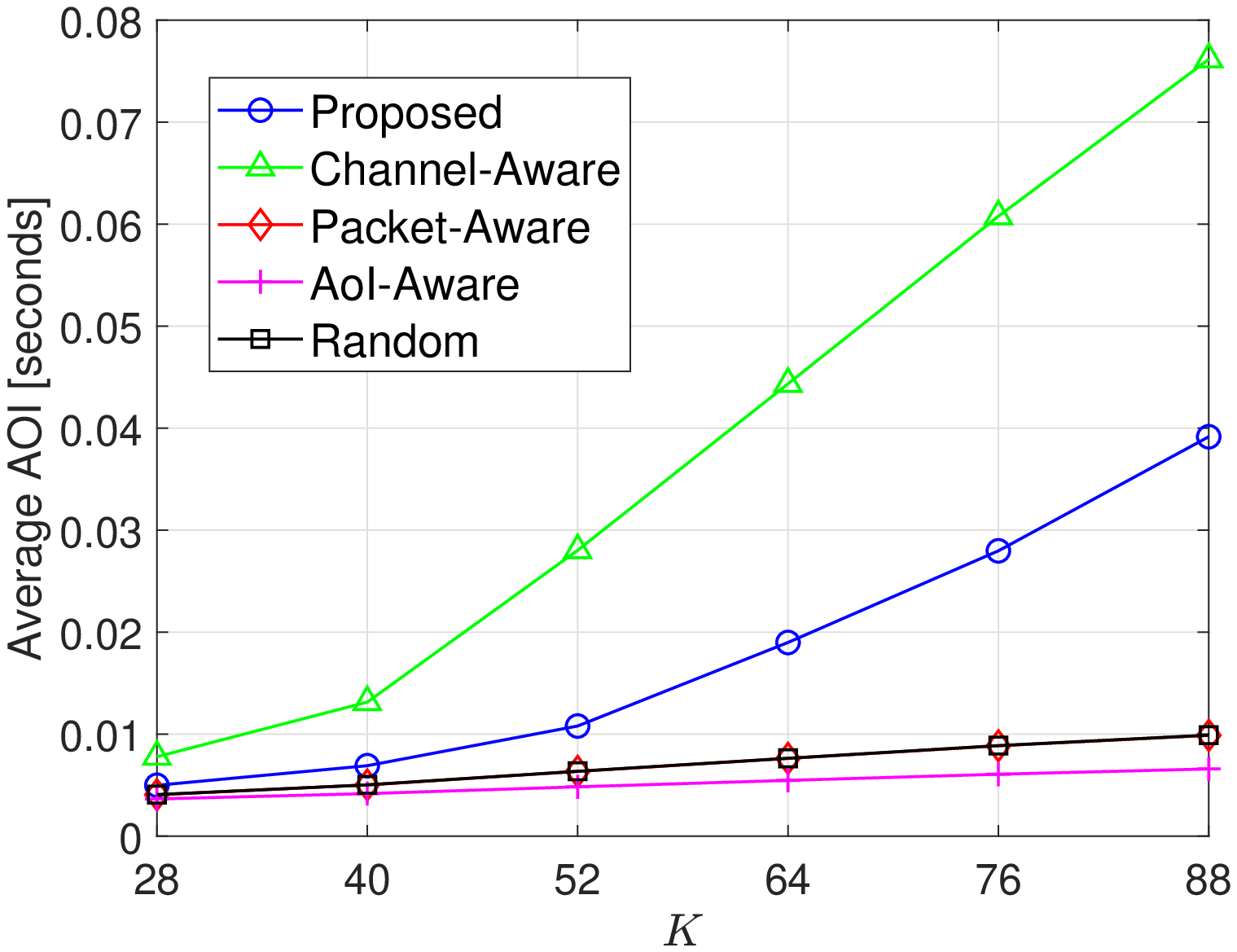}
     \caption{Average AoI per VUE-pair across the time horizon versus $K$: $B = 3$, $\ell = 65$ m and $\lambda = 6$ packets/slot.}
     \label{sim04_03}
   \end{minipage}
   \hspace{0.1cm}
   \begin{minipage}{0.48\textwidth}
     \centering
     \vspace{1cm}
     \includegraphics[width=\textwidth]{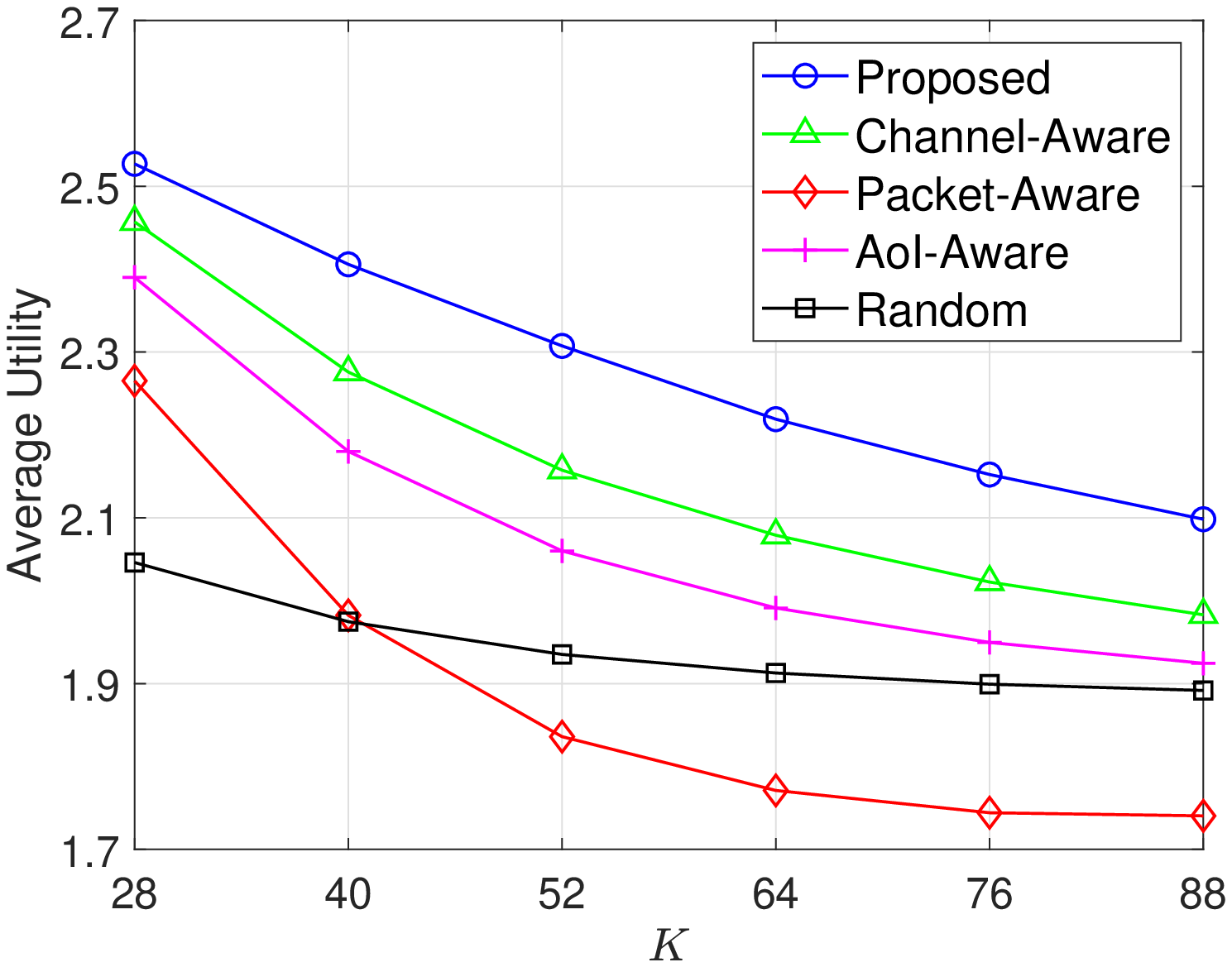}
     \caption{Average utility per VUE-pair across the time horizon versus $K$: $B = 3$, $\ell = 65$ m and $\lambda = 6$ packets/slot.}
     \label{sim04_04}
   \end{minipage}
\end{figure}





The last experiment compares the average performance from the proposed proactive algorithm to the other four baseline algorithms versus different numbers of VUE-pairs.
We assume a Manhattan grid V2V communication network in which the VUE-pair distance and the average packet arrival rate are $\ell = 65$ m and $\lambda = 6$ packets/slot.
Figs. \ref{sim04_01}, \ref{sim04_02}, \ref{sim04_03} and \ref{sim04_04} draw the curves of average transmit power consumption, average packet drops, average AoI and average utility per VUE-pair over the scheduling slots.

It is straightforward that more VUE-pairs means less opportunities to be allocated a frequency band for the transmission of scheduled data packets, which indicates for all algorithms, smaller average transmit power consumption (as shown in Fig. \ref{sim04_01}), more average packet drops (as shown in Fig. \ref{sim04_02}), larger average AoI (as shown in Fig. \ref{sim04_03}), and hence with the current weight value settings, worse utility performance (as shown in Fig. \ref{sim04_04}).
Note that in last three experiments, the Queue-Aware and the Random baseline algorithms achieve almost the same average AoI performance.
The reason behind this observation is that with the Queue-Aware baseline, the RSU allocate as with the Random baseline the frequency bands according to the randomly generated variables, which are the packet arrivals.
All these experiments clearly illustrate that the proposed proactive algorithm is able to ensure better average utility performance for the VUE-pairs than the other four baseline algorithms.

\section{Conclusions}
\label{conc}

In this paper, we investigate the AoI-aware RRM problem for an expected long-term performance optimization in a Manhattan grid V2V communication network.
The RSU allocates frequency bands and schedules data packet transmissions for all VUE-pairs according to the observations of global network states across the discrete time horizon.
The decision-making process falls into the realm of a single-agent MDP.
The daunting technical challenges in solving an optimal control policy prompt us to first decompose the MDP at the RSU into the per-VUE-pair MDPs with much simplified decision makings.
Furthermore, to overcome the partial observability and the curse of high dimensionality in local network state space of a VUE-pair, we resort to the LSTM technique and the DQN, and propose a proactive algorithm based on the DRQN.
%
%
The proposed algorithm enables frequency band allocation and packet scheduling decisions with only local partial network state observations from the VUE-pairs but without a priori statistics knowledge of network dynamics.
Numerical experiments show significant gains in average utility performance from the proposed proactive algorithm compared with four other state-of-the-art baselines.

\end{document}